\documentclass[lettersize,journal]{IEEEtran}

\usepackage{amsfonts,amssymb,amsmath}
\usepackage{hyperref}
\usepackage{graphicx}
\usepackage{booktabs}
\usepackage{embrac}

\usepackage{subeqnarray}
\usepackage{cases}
\usepackage{enumerate}
\usepackage{multirow}
\usepackage{multicol}
\usepackage{makecell}
\usepackage{algpseudocode}

\usepackage{algorithm}
\usepackage{array}
\usepackage{textcomp}
\usepackage{stfloats}
\usepackage{url}
\usepackage{verbatim}
\usepackage{cite}
\usepackage{soul}
\usepackage{color, xcolor}
\soulregister{\cite}7
\soulregister{\ref}7

\usepackage{float}
\usepackage{subcaption}
\usepackage{hyperref}
\hypersetup{hidelinks,
        	colorlinks=true,
        	allcolors=blue,
        	pdfstartview=Fit,
        	breaklinks=true}

\begin{document}

\title{A Relay System for Semantic Image Transmission based on Shared Feature Extraction and Hyperprior Entropy Compression}

\author{Wannian An, Zhicheng Bao, Haotai Liang, Chen Dong*, and Xiaodong Xu,~\IEEEmembership{Senior Member,~IEEE}
\thanks{This work was supported by the National Key Research and Development Program of China under Grant 2022YFB2902102.
\emph{(Corresponding author: Chen Dong.)}

Wannian An, Zhicheng Bao, Haotai Liang, Chen Dong and Xiaodong Xu are with the State Key Laboratory of Networking and Switching Technology, Beijing University of Posts and Telecommunications, Beijing, 100876, China. (e-mail: anwannian2021@bupt.edu.cn; zhicheng\_bao@bupt.edu.cn; lianghaotai@bupt.edu.cn; dongchen@bupt.edu.cn; xuxiaodong@bupt.edu.cn;).}}

\markboth{Journal of \LaTeX\ Class Files,~Vol.~XX, No.~XX, XXX~2023}
{Shell \MakeLowercase{\textit{et al.}}: Opportunistic Routing Aided Cooperative Communication Network with Energy Harvesting}

\maketitle

\begin{abstract}
Nowadays, the need for high-quality image reconstruction and restoration is more and more urgent. However, most image transmission systems may suffer from image quality degradation or transmission interruption in the face of interference such as channel noise and link fading. To solve this problem, a relay communication network for semantic image transmission based on shared feature extraction and hyperprior entropy compression (HEC) is proposed, where the shared feature extraction technology based on Pearson correlation is proposed to eliminate partial shared feature of extracted semantic latent feature. In addition, the HEC technology is used to resist the effect of channel noise and link fading and carried out respectively at the source node and the relay node. Experimental results demonstrate that compared with other recent research methods, the proposed system has lower transmission overhead and higher semantic image transmission performance. Particularly, under the same conditions, the multi-scale structural similarity (MS-SSIM) of this system is superior to the comparison method by approximately 0.2.
\end{abstract}

\begin{IEEEkeywords}
Semantic communication, relay system, shared feature extraction, hyperprior entropy compression.
\end{IEEEkeywords}

\IEEEpeerreviewmaketitle

\section{Introduction}
\IEEEPARstart{S}{emantic} communication technology is considered to be one of the key technologies of future mobile communication \cite{1,2}. According to Shannon and Weaver's information theory \cite{3}, semantic communication is located at the second level of the three levels of information transmission. The goal of semantic communication is to accurately transmit semantic information in the original data, rather than accurately transmit the bit information of the original data.

For data with different structures, such as text \cite{4,5}, image \cite{5,6}, voice \cite{7} and video \cite{8,29}, semantic information is processed in different ways. Especially, literature \cite{4} develops a text semantic communication system named DeepSC based on the text Tansformer framework. The layer-based semantic communication system for image (LSCI) is proposed in \cite{6} to realize image semantic extraction and reconstruction. In literature \cite{7}, a squeeze-and-excitation network is used to develop a semantic communication system named DeepSC-S, which is based on the attentional mechanism for the transmission of voice signals. In addition, a semantic video conferencing network based on key-point transmission is proposed \cite{8}, where an incremental redundancy hybrid automatic repeat-request framework based on semantic error detector is developed. Actually, compared with text modal and speech modal, the visual modal encompasses a wealth of information, including rich information such as color, shape, texture, etc. With the rapid increase of the demand for high-quality transmission of image signals \cite{9}, the research of semantic image transmission gradually become a hot spot in semantic communication research.


A deep joint source channel coding (JSCC) technology for wireless image transmission is proposed \cite{13}, which directly maps image pixel values to complex channel input symbols, and verifies that the JSCC technology is not affected by cliff effect. A practical multi-description JSCC scheme is proposed in \cite{14} for adaptive bandwidth image transmission over wireless channels. Additionally, the attention deep learning based JSCC scheme is proposed in \cite{15}, which employs channel-wise soft attention to adjust feature scaling based on signal-to-noise ratio (SNR) conditions. In \cite{16}, the heterogeneous communication framework is studied, where semantic communication and traditional communication coexist. The non-orthogonal multiple access -based multi-user semantic communication (NOMASC) system is proposed in \cite{28} to support the semantic transmission of multiple users.

Compared with the aforementioned deep JSCC schemes \cite{13,14,15,16,28}, the codec schemes that combine deep JSCC with feature importance (FI) have better performance in image processing. In literature \cite{17}, the semantic transmission of aerial image based on unmanned aerial vehicle is studied, which achieves the balance between uplink transmission delay and classification accuracy, using the nonlinear transformation of block selection and compression of feature information. A shared features extraction technology based on the distance of feature elements is proposed to extract the shared feature redundancy in image semantic features \cite{18}, so as to reduce the transmission bandwidth of semantic information. A nonlinear transform source-channel coding (NTSCC) for image semantic transmission is proposed \cite{19}, and the essence is to learn the hyperprior entropy model (HEM) of potential representation of source data, so that to implicitly approximate the real source distribution. Based on this entropy model, an adaptive rate transmission and hyperprior assisted encoding and decoding mechanism is designed to improve the performance of the classical deep JSCC. In literature \cite{20}, a deep video semantic transmission (DVST) framework is studied on the basis of literature \cite{19}, where nonlinear transformation and conditional coding architecture are used to adaptively extract semantic features between video frames. Compared with traditional wireless video encoding transmission schemes, the proposed DVST has better transmission performance.

It is worth noting that all the semantic communication systems described above are end-to-end (E2E) communication systems. However, relay communication plays an important role in resisting channel fading and expanding signal coverage \cite{21,22,27}. Different from traditional relays, semantic relays ensure accurate forwarding of semantic information rather than bit information \cite{23,24}. In this paper, we investigate a relay communication network for semantic image transmission. In the process of semantic image transmission, the shared feature extraction technology based on Pearson correlation is used to eliminate partial shared latent features, and the hyperprior entropy compression (HEC) technology is used to effectively compress transmission data under the condition of channel noise and link fading. Table~\ref{I} shows the comparison between our work and the above references. The main contributions of this paper are summarized as follows:
\begin{enumerate}
  \item \emph{Twice Compressed Semantic Image Relay Network:} In this paper, the twice compressed semantic image relay network is proposed, where the semantic features transmitted are compressed by the HEC technology according to the condition of channel noise and link fading.
  \item \emph{Shared Feature Extraction Technology based on Pearson Correlation:} In order to effectively reduce the semantic latent feature space dimension in the transmission process, the shared feature extraction technology based on Pearson correlation is proposed, which makes the encoding and transmission of semantic information more efficient.
  \item \emph{Performance Verification:} The effectiveness of the proposed semantic image relay communication system was verified by comparing it with other recent research methods, such as the shared extraction technology based on the distance of semantic feature elements. In particular, under the same conditions, the proposed system can achieve an MS-SSIM advantage of about 0.2 compared with the comparison method.
\end{enumerate}

\begin{table}[!t]
\centering
\caption{Comparison of references. Where N indicates that the technology is not adopted in the study, and Y indicates that the technology is adopted in the study.}
\label{I}
\renewcommand{\arraystretch}{1.3}
\begin{tabular}{|c|c|c|c|c|}
\hline
Reference & JSCC & JSCC+FI & System\\
\hline
\begin{tabular}[c]{@{}c@{}}\cite{13,14,15,16,28} \end{tabular} & Y & N & E2E\\
\hline
\begin{tabular}[c]{@{}c@{}}\cite{17,18,19,20} \end{tabular} & N & Y & E2E\\
\hline
\begin{tabular}[c]{@{}c@{}}\cite{23}, \cite{24} \end{tabular} & Y & N & Relay\\
\hline
This work & N & Y & Relay\\
\hline
\end{tabular}
\end{table}

The remainder of this paper is organized as follows. In section II, the system model is described. In section III, the proposed data processing methods are shown. The numerical results are presented in Section IV. Finally, the conclusion is presented in Section V.

$Notations:$ $A\sim\mathcal{CN}(0, \theta)$ indicates the random variable $A$ follows the complex Gaussian distribution with mean 0 and variance $\theta$. $\left \lfloor \cdot \right \rfloor$ indicates round-down operation. $\left \lfloor \cdot \right \rceil$ represents the operation of rounding to an integer. The absolute value of $B$ is denoted by $|B|$. $\widetilde{Y}$ is the quantized representation of $Y$. $\mathbb{E}[\cdot]$ denotes the expectation operator. Boldface capital and lower-case letters stand for matrices and vectors, respectively. $\mathbf{e}$ denotes identity vector and $\mathbf{E}$ denotes identity matrix. $\mathbb{R}^k$ means an $k$-dimensional real number field space. $d(a,b)$ indicates the mean squared error (MSE) between $a$ and $b$. $\binom{n}{k}$ represents the number of combinations of selecting $k$ elements from $n$ elements.

\section{System Model}

\subsection{Overall Architecture}
\begin{figure*}
\centerline{\includegraphics[width=7in]{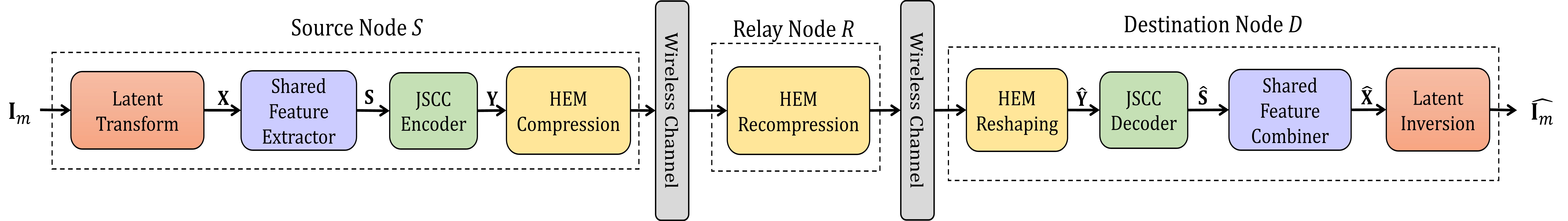}}
\caption{System model architecture, which consists a source node $S$ responsible for semantic extraction, a relay node $R$ responsible for semantic forwarding, and a destination node $D$ responsible for semantic recovery. In each transmission cycle, the source node $S$ extracts semantic information from input images and transmits it to the relay node $R$. Subsequently, the relay node $R$ forwards the received semantic information to the destination node $D$. \label{fig1}}

\end{figure*}

The overall architecture of the semantic image transmission relay system is depicted in Fig. \ref{fig1}, comprising three essential components: a source node denoted as $S$, a relay node denoted as $R$, and a destination node denoted as $D$. The source node $S$ is primarily responsible for semantic extraction, while the relay node $R$ facilitates semantic forwarding, and the destination node $D$ handles semantic recovery. Within this communication system, it is assumed that both the source node $S$ and the destination node $D$ possess an identical background knowledge base. Furthermore, the transmit power $P$ provided by both the source node $S$ and the relay node $R$ remains constant regardless of the volume of data being transmitted. The subsequent subsections will provide detailed descriptions of the functions of each component in the semantic image transmission relay system.


\subsection{Model Design for the Source Node $S$}
The source node $S$ consists of four primary modules: the latent transform module, the shared feature extractor module, the JSCC-encoder module, and the HEM compression module. These modules collectively serve the purposes of information preprocessing, latent space merging, joint source-channel coding, and compression of coded semantic information prior to transmission. The specific details of each module are as follows:

\subsubsection{latent transform module}
As illustrated in Fig. \ref{fig1}, the source node $S$ employs a latent transform module, which is mainly implemented by convolutional neural network structure. This module aims to convert the input RGB image data $\mathbf{I}_m = \{\mathbf{I}_{m1}, \mathbf{I}_{m2}, \dots, \mathbf{I}_{mN}\}$ into a low-dimensional latent feature space, facilitating more effective extraction of semantic features. The transformation process is expressed as follows:
\begin{equation}\label{eq1}
\mathbf{X}_i = LT_{e}(\mathbf{I}_{mi}, \alpha_{e}),\, i\in\{1,2,\dots,N\},
\end{equation}
where $LT_{e}(\cdot)$ denotes the latent transform operation with parameter $\alpha_{e}$. $\mathbf{I}_{mi}\in \mathbb{R}^{W\times H \times 3}$ represents the $i$-th input image data, consisting of a width $W$, height $H$, and RGB three channels. Moreover, $\mathbf{X}_{i} \in \mathbb{R}^{W\times H \times C}$ represents the latent feature space corresponding to $\mathbf{I}_{mi}$, with dimensions of width $W$, height $H$, and $C$ channels.

\subsubsection{shared feature extractor module}
Since the input images of the source node $S$ are sourced from the same background knowledge base, the generated latent feature space $\mathbf{X} =\{\mathbf{X}_{1}, \mathbf{X}_{2}, \dots, \mathbf{X}_{i}, \dots, \mathbf{X}_{N}\}$ exhibits certain similarities. To further simplify the latent feature space $\mathbf{X}$ and reduce the complexity of semantic feature coding based on it, a shared feature extractor based on Pearson correlation is employed. This extractor diminishes redundant shared features within the latent feature space $\mathbf{X}$. The merged latent feature space, denoted as $\mathbf{S} \in \mathbb{R}^{W \times H \times C2}$, contains multiple input image information. The merging process is represented as follows:
\begin{equation}\label{eq2}
            \mathbf{S} = SM_{p}(\mathbf{X}, \gamma_{p}),
\end{equation}
where $SM_{p}(\cdot)$ denotes the merging process with a  shared information extraction rate parameter $\gamma_{p}$. For further details on the merging process, please refer to subsection III-A.

\subsubsection{JSCC-encoder module}
In order to effectively resist the influence of channel fading during $SR$ link transmission, JSCC of latent feature space $\mathbf{S}$ is carried out as follows:
\begin{equation}\label{eq3}
            \mathbf{Y} = A_{e}(\mathbf{S},\varphi_{e}),
\end{equation}
where $\mathbf{Y}\in \mathbb{R}^{W \times H \times C2}$ denotes the encoded semantic feature data, while $A_{e}(\cdot)$ represents the encoder, which consists of a multi-layer convolutional structure and takes $\varphi_{e}$ as the parameter.

\subsubsection{HEM compression module}
To mitigate the impact of channel fading during $SR$ link transmission, the coding semantic feature compression based on HEC technology is employed. This approach selects a subset of the encoded semantic feature data $\mathbf{Y}$ for transmission, taking into account the importance $\mathbf{I}\in \mathbb{R}^{W \times H \times C2}$ of the encoded semantic feature data and the specified compression rate $v1$ of the source node $S$. The process of obtaining the compressed semantic feature data $\mathbf{S1}\in \mathbb{R}^{1\times K1}$ is represented as follows:
\begin{equation}\label{eq4}
\mathbf{S1} = C_{1}(\mathbf{Y},\mathbf{I}, v1),
\end{equation}
where $C_{1}(\cdot)$ denotes a compression transformation with the compression ratio $v1 \in (0, 1)$ as the parameter. For a detailed explanation of the compression process, please refer to subsection III-B-1.


After power normalization, the compressed feature $\mathbf{S1}$ is transmitted over the wireless channel to the relay node $R$. The received semantic feature data $\widehat{\mathbf{S1}}\in \mathbb{R}^{1\times K1}$ at the relay node $R$ is expressed as:
\begin{equation}\label{eq5}
\widehat{s1} = \sqrt{\bar{P}}{h}_{SR}s1+{n}_{R}.
\end{equation}

In the equation above, $\bar{P}$ represents the average transmit power for each semantic feature data. $\widehat{s1}$ and $s1$ are the elements of $\widehat{\mathbf{S1}}$ and $\mathbf{S1}$ respectively. $h_{SR}\sim \mathcal{N}(0,d_{SR}^{-a})$ denotes the Rayleigh fading channel between the $SR$ link, which remains constant over a transmission period. $n_{R}\sim \mathcal{N}(0,N_R)$ represents the AWGN at the relay node $R$. Specifically, $d_{SR}$ is the distance between the source node $S$ and the relay node $R$, $a$ is the path-loss parameter, and $N_R$ represents the power of the noise received at the relay node $R$.


\subsection{Model Design for the Relay Node $R$}
The relay node $R$ incorporates a HEM recompression module, which is primarily responsible for recompression of the received semantic feature data based on the condition of fading in the $RD$ link. Upon receiving the semantic feature data $\widehat{\mathbf{S1}}$ from the source node $S$, along with the corresponding importance information $\mathbf{I}$ for the semantic feature data $\widehat{\mathbf{S1}}$, the relay node $R$ chooses a portion of the received semantic feature data $\widehat{\mathbf{S1}}$ based on the importance information $\mathbf{I}$. Subsequently, the relay node $R$ transmits the selected data to the destination node $D$. The process of obtaining the compressed semantic feature data $\mathbf{S2}\in \mathbb{R}^{1\times K2}$ at the relay node $R$ is expressed as follows:
\begin{equation}\label{eq6}
\begin{split}
\widehat{\mathbf{Y1}} &= C_1^{-1}(\widehat{\mathbf{S1}}, \mathbf{I}),\\
\mathbf{S2} &= C_2(\widehat{\mathbf{Y1}}, \mathbf{I}, v2),
\end{split}
\end{equation}
where $C_1^{-1}$ represents the reshaping transformation, and $C_2(\cdot)$ denotes a compression transformation with the compression ratio $v2 \in(0,1)$ as the parameter. For a more detailed explanation of the recompression process, please refer to subsection III-B-2.


After power normalization, the compressed feature $\mathbf{S2}$ is transmitted over the wireless channel to the destination node $D$. The received signal $\widehat{\mathbf{S2}}\in \mathbb{R}^{1\times K2}$ at the destination node $D$ is expressed as:

\begin{equation}\label{eq7}
\widehat{s2} = \sqrt{\bar{P}}h_{RD}s2+n_{D}.
\end{equation}

In the equation above, $\widehat{s2}$ and $s2$ represent the elements of $\widehat{\mathbf{S2}}$ and $\mathbf{S2}$ respectively. $h_{RD}\sim \mathcal{N}(0,d_{RD}^{-a})$ denotes the Rayleigh fading channel between the $RD$ link, which remains constant over the transmission period. $n_{D}\sim \mathcal{N}(0,N_D)$ represents the AWGN at the destination node $D$. Specifically, $d_{RD}$ refers to the distance between the relay node $R$ and the destination node $D$, and $N_D$ represents the power of the noise received at the destination node $D$.


\subsection{Model Design for the Destination Node $D$}
The destination node $D$ consists of four main modules: HEM reshaping, JSCC-decoder, shared feature combiner, and latent inversion. These modules are responsible for performing sparse reshaping of the received semantic feature data, joint source channel decoding, splitting the latent space, and recovering the semantic features. The details of each module are as follows:

\subsubsection{HEM reshaping module}
Upon receiving the semantic feature data $\widehat{\mathbf{S2}}$ and the corresponding importance information $\mathbf{I}$ from the relay node $R$, the destination node $D$ performs a sparsely reshaping operation on the received semantic feature data. This reshaping is carried out based on the importance information to recover the spatial location information of the transmitted semantic feature data. The detailed process of reshaping is described in section III-B-3. The reshaped semantic feature data $\widehat{\mathbf{Y}}\in \mathbb{R}^{W \times H \times C2}$ is expressed as follows:

\begin{equation}\label{eq8}
            \widehat{\mathbf{Y}} =  C_2^{-1}(\widehat{\mathbf{S2}}, \mathbf{I}),
\end{equation}
where $C_2^{-1}(\cdot)$ represents a reshaping transformation.
\subsubsection{JSCC-decoder module}
This module conducts joint source channel decoding on the input sparse semantic feature data, aiming to map it back to the approximate space of the merged latent feature representation in the source node $S$. The decoding process is described as follows:
\begin{equation}\label{eq9}
\widehat{\mathbf{S}} = A_{d}(\widehat{\mathbf{Y}},\theta_{d}),
\end{equation}
where $\widehat{\mathbf{S}}\in \mathbb{R}^{W \times H \times C2}$ represents the obtained latent feature space after decoding. The decoding operation is performed by an decoder $A_{d}(\cdot)$, which consists of a multi-layer convolutional structure and takes $\theta_{d}$ as the parameter.


\subsubsection{shared feature combiner module}
The latent feature space $\widehat{\mathbf{S}}$ contains the latent features corresponding to the transmitted multiple images. In order to effectively recover the information of the transmitted multiple images, it is necessary to separate the latent feature space $\widehat{\mathbf{X}} =\{\widehat{\mathbf{X}_{1}}, \widehat{\mathbf{X}_{2}}, \dots, \widehat{\mathbf{X}_{i}}, \dots, \widehat{\mathbf{X}_{N}}\}$ corresponding to each transmitted image from $\widehat{\mathbf{S}}$. The detailed process of separation is described in subsection III-A.
\subsubsection{latent inversion module}
The latent inversion module consists of a multi-layer transposed convolutional network designed to map the latent feature space back to the original RGB image data, thereby completing the semantic transmission of images. The transformation process is expressed as follows:
\begin{equation}\label{eq10}
            \widehat{\mathbf{I}_{mi}} = LT_{d}(\widehat{\mathbf{X}_{i}}, \alpha_{d}),\, i\in\{1,2,\dots,N\}.
\end{equation}
where $LT_{d}(\cdot)$ represents the latent inversion transform parameterized by $\alpha_{d}$. The output $\widehat{\mathbf{I}_{mi}}\in \mathbb{R}^{W\times H \times 3}$ represents the $i$-th reconstructed image data.

\section{The Proposed Data Processing Methods}
\begin{figure}
\centerline{\includegraphics[width=3.5in]{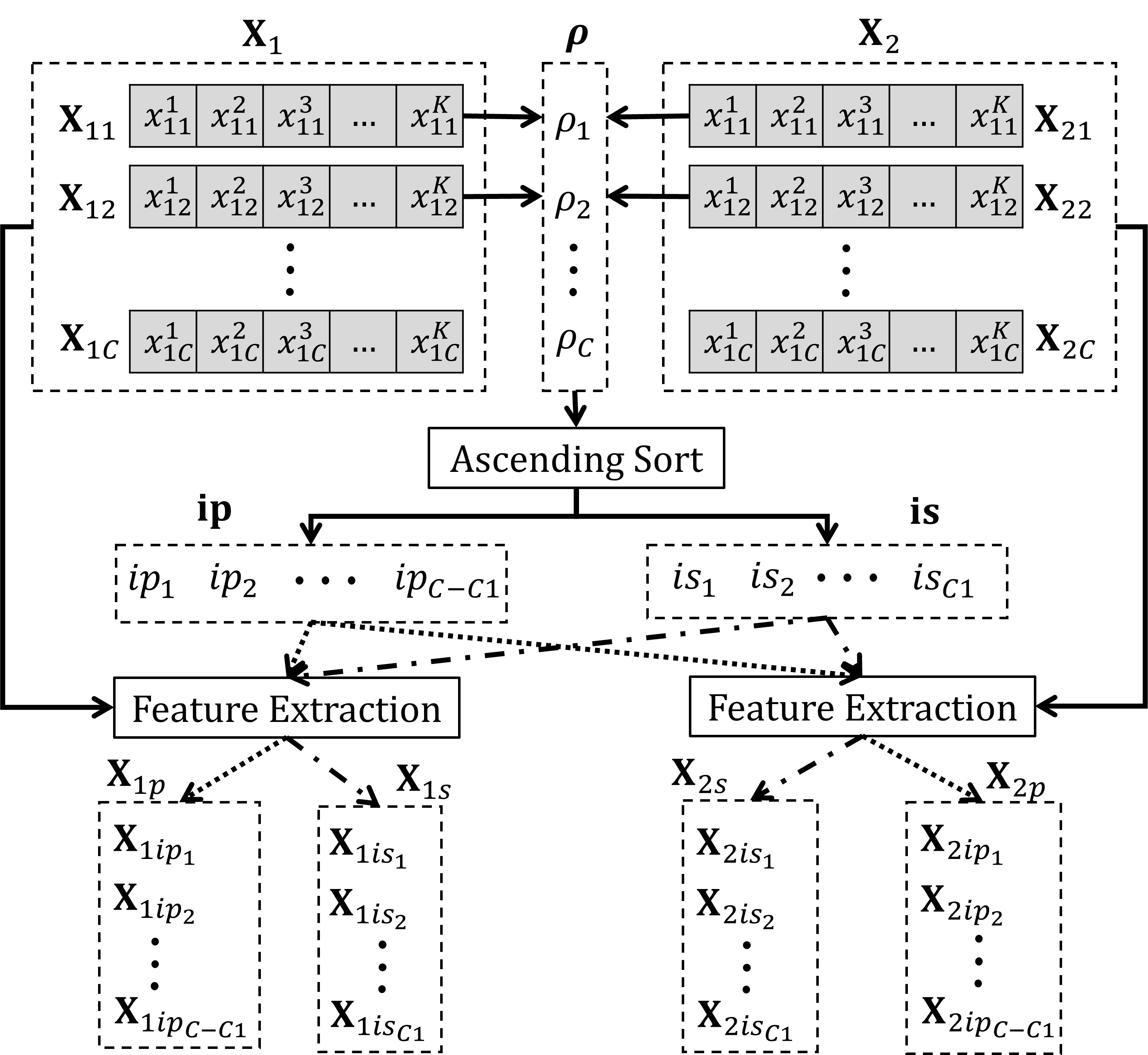}}
\caption{Shared feature extraction technology based on Pearson correlation for two latent feature spaces, which is used to partition the latent feature space $\mathbf{X}_{i}(i\in {1,2})$ into personalized latent feature subspace $\mathbf{X}_{ip}$ and shared latent feature subspace $\mathbf{X}_{is}$.\label{fig2}}
\end{figure}
\subsection{Shared Feature Extraction Technology}
In order to effectively reduce the information redundancy in latent feature space $\mathbf{X} =\{\mathbf{X}_{1}, \mathbf{X}_{2}, \dots, \mathbf{X}_{i}, \dots, \mathbf{X}_{N}\}$, a shared feature extraction technology based on Pearson correlation is employed to partition the latent feature space $\mathbf{X}_{i}$ into personalized latent feature subspace $\mathbf{X}_{ip}$ and shared latent feature subspace $\mathbf{X}_{is}$. Fig. \ref{fig2} illustrates the partitioning process for the case of $N=2$, and the specific partitioning process is as follows:
\begin{itemize}
\item Similarity measurement of output channel features for latent transform module: Calculate the Pearson correlation coefficient $\rho_{c} (c\in\{1,2,\ldots,C\})$ between the feature vectors $\mathbf{x}_{1c}\in \mathbb{R}^{1\times K}$ and $\mathbf{x}_{2c}\in \mathbb{R}^{1\times K}$, which are obtained by flattening the feature matrices in the latent feature spaces $\mathbf{X}_{1}$ and $\mathbf{X}_{2}$, where $K=WH$. The Pearson correlation coefficient $\rho_{c}$ is computed as follows:
\begin{equation}\label{eq11}
            \rho_{c} = \left| \frac{\sum_{k=1}^K (x_{1c}^k-\mu_1)(x_{2c}^k-\mu_2)}{\sqrt{\sum_{k=1}^K (x_{1c}^k-\mu_1)^2}\sqrt{\sum_{k=1}^K (x_{2c}^k-\mu_2)^2}} \right|,
\end{equation}
where $\mu_1$ and $\mu_2$ represent the means of the vectors $\mathbf{x}_{1c}$ and $\mathbf{x}_{2c}$, respectively. By utilizing Eq. (\ref{eq11}), the Pearson correlation coefficient vector $\boldsymbol{\rho} = [\rho_{1},\rho_{2},\ldots,\rho_{C}]$ corresponding to the features $\mathbf{X}_{1}$ and $\mathbf{X}_{2}$ can be calculated.

\item Partition of personalized latent feature subspace and shared latent feature subspace: To begin, set the shared information extraction rate $\gamma_{p} \in(0,1)$, sort the elements of $\boldsymbol{\rho}$ in ascending order. Next, create the shared channel index vector $\mathbf{is}$ by selecting the indices of the $C1 = \left \lfloor \gamma_{p} C \right \rfloor$ larger elements in $\boldsymbol{\rho}$.
    Additionally, the indices of the remaining elements in $\boldsymbol{\rho}$ form the personalized channel index vector $\mathbf{ip}$. Finally, extract the shared latent feature subspace $\mathbf{X}_{is}\in \mathbb{R}^{W\times H \times C1}$ and the personalized latent feature subspace $\mathbf{X}_{ip}\in \mathbb{R}^{W\times H \times (C-C1)}$ from $\mathbf{X}_{i} , (i \in{1,2})$, which performed based on the index vectors $\mathbf{is}$ and $\mathbf{ip}$, respectively.
\end{itemize}

Furthermore, in the case of $N>2$, the partitioning process differs from the case of $N=2$ in the following manner: Firstly, the Pearson correlation coefficient vector $\boldsymbol{\rho}_j \, (j\in\{1,2,\dots, \binom{N}{2}\})$ is calculated between any two semantic features $\mathbf{X}_{i1}$ and $\mathbf{X}_{i2}\, (i1\neq i2, i1,i2\in\{1,2,\dots,N\})$. Then, the minimum value of the corresponding element at each position in all $\boldsymbol{\rho}_j$ vectors is selected to form the Pearson correlation coefficient vector $\boldsymbol{\rho}$. This operation ensures the establishment of a lower bound for the correlation among all semantic features.


In order to facilitate the sharing of the shared latent feature subspace and the personalized latent feature subspace between the source node $S$ and the destination node $D$, a merging protocol is proposed. This protocol combines the personalized latent feature subspace $\mathbf{X}_{ip},(i\in{1,2,\dots, N})$ and the shared latent feature subspace $\mathbf{X}_{s}$ along the channel dimension. The merged latent information space $\mathbf{S} \in \mathbb{R}^{W \times H \times C2}$, where $C2 = N(C-C1)+C1$, is then transmitted. The specific execution process of this protocol is illustrated as follows:
\begin{equation}\label{eq12}
        \mathbf{S} = cat\Big((\mathbf{X}_{1p}, \mathbf{X}_{s}, \mathbf{X}_{2p}, \dots, \mathbf{X}_{Np}), dim=channel \Big),
\end{equation}
where $cat\big((\cdot), dim=channel\big)$ indicates the features merging operation in the channel dimension. Moreover, the shared latent information subspace $\mathbf{X}_{s}$ is obtained as follows
\begin{equation}\label{eq13}
            \mathbf{X}_{s} = ave\Big(\mathbf{X}_{1s},\mathbf{X}_{2s},\dots, \mathbf{X}_{Ns}),dim=channel \Big),
\end{equation}
where $ave\big((\cdot), dim=channel\big)$ indicates the operation of calculating the feature average value in the channel dimension.

After decoding the merged latent feature space $\widehat{\mathbf{S}}$, the destination node $D$ identifies the personalized latent information subspace $\widehat{\mathbf{X}_{ip}}$ and the shared latent information subspace $\widehat{\mathbf{X}_{is}}$ for each image based on the merging protocol, the shared information extraction rate $\gamma_{p}$ and the number $C$ of channels. Subsequently, the destination node $D$ performs a channel-wise combination operation on $\widehat{\mathbf{X}_{ip}}$ and $\widehat{\mathbf{X}_{is}}$, resulting in the latent information space $\widehat{\mathbf{X}_{i}}$ corresponding to each image. The process is demonstrated as follows:

\begin{equation}\label{eq14}
            \widehat{\mathbf{X}_{i}} = cat\Big(\widehat{\mathbf{X}_{ip}},\widehat{\mathbf{X}_{is}}),dim=channel \Big).
\end{equation}

From the aforementioned combination process, it can be concluded that there is no need for additional transmission of the partitioned shared channel index vector $\mathbf{is}$ between the source node $S$ and the destination node $D$.
\subsection{HEC Technology}
\begin{figure*}
\centerline{\includegraphics[width=7in]{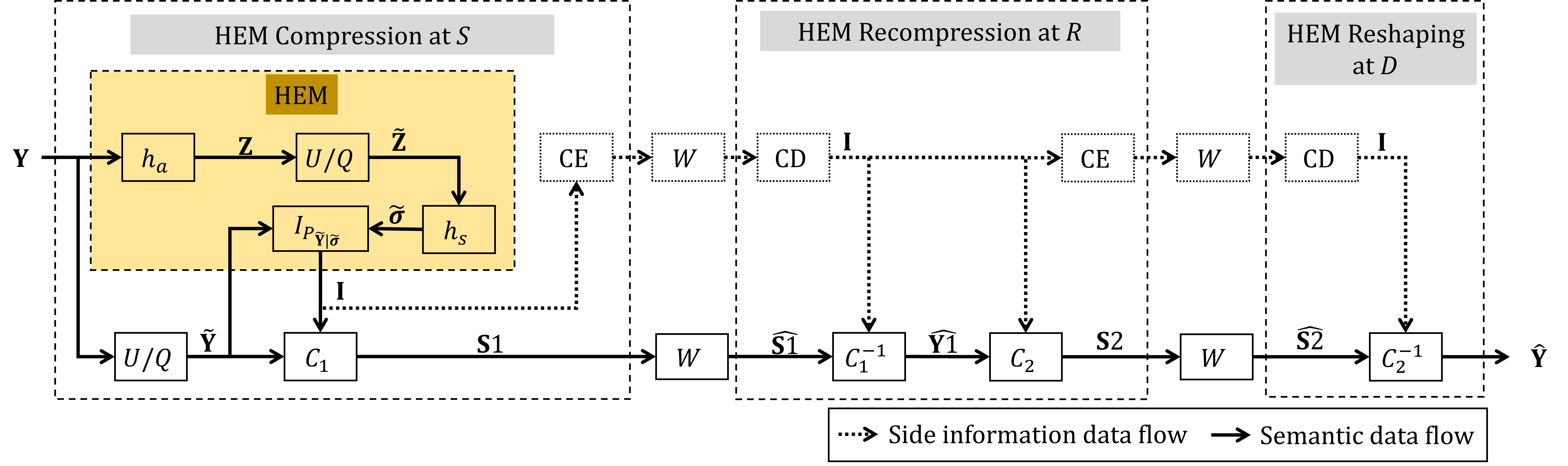}}
\caption{Double compression of semantic feature data based on the HEM in the $SR$ link transmission process and the $RD$ link transmission process. Boxes denote data transformation or quantization, arrows represent the flow of data, $W$ denotes the wireless channel, CE represents the  channel encoding and CD denotes the channel decoding. Additionally, $U$ indicates the addition of uniform noise during model training, while $Q$ denotes the application of uniform scalar quantization $\left \lfloor \cdot \right \rceil$ (rounding to integers) during model testing. \label{fig3}}
\end{figure*}
\subsubsection{HEC technology used at the source node $S$}
As depicted in Fig. \ref{fig3}, in order to ensure efficient transmission in the fading channel, HEC technology is employed to compress the encoded feature $\mathbf{Y}$ at the source node $S$. Additionally, it is necessary to obtain the entropy model $\mathbf{P}_{\widetilde{\mathbf{Y}}|\widetilde{\boldsymbol{\sigma}}}$ of the encoded feature $\mathbf{Y}$ for effective compression of $\mathbf{Y}$. According to \cite{25}, the $j$-th element $\widetilde{y}_{j}$ of the quantized encoded feature $\widetilde{\mathbf{Y}}$ can be modeled as a random variable following the Gaussian distribution $\mathcal{N}(0,\widetilde{\sigma}_{j}^2)$ as follows:
\begin{equation}\label{eq15}
            P_{\widetilde{y}_{j}|\widetilde{\sigma}_{j}} = \Big(\mathcal{N}(0,\widetilde{\sigma}_{j}^2)*\mathcal{U}(-\frac{1}{2},\frac{1}{2}) \Big)(\widetilde{y}_{j}),
\end{equation}
where $*$ denotes the convolutional operation. Furthermore, the standard deviation parameter $\widetilde{\sigma}$ for all elements of $\widetilde{\mathbf{Y}}$ is obtained by the following nonlinear transformation:
\begin{equation}\label{eq16}
            \widetilde{\boldsymbol{\sigma}} = h_{s}(\widetilde{\mathbf{Z}},\theta_{h}),
\end{equation}
where $h_{s}(\cdot)$ is a nonlinear transformation parameterized by $\theta_{h}$, $\widetilde{\mathbf{Z}}$ represents the quantized hyperprior information $\mathbf{Z}$ of feature $\mathbf{Y}$, and the extraction process of the hyperprior information $\mathbf{Z}$ is defined as follows:
\begin{equation}\label{eq17}
            \mathbf{Z} = h_{a}(\mathbf{Y},\varphi_{h}),
\end{equation}
where the nonlinear transformation $h_{a}(\cdot)$ is a feature compressor parameterized by $\varphi_{h}$.

\begin{figure*}
\centerline{\includegraphics[width=6.6in]{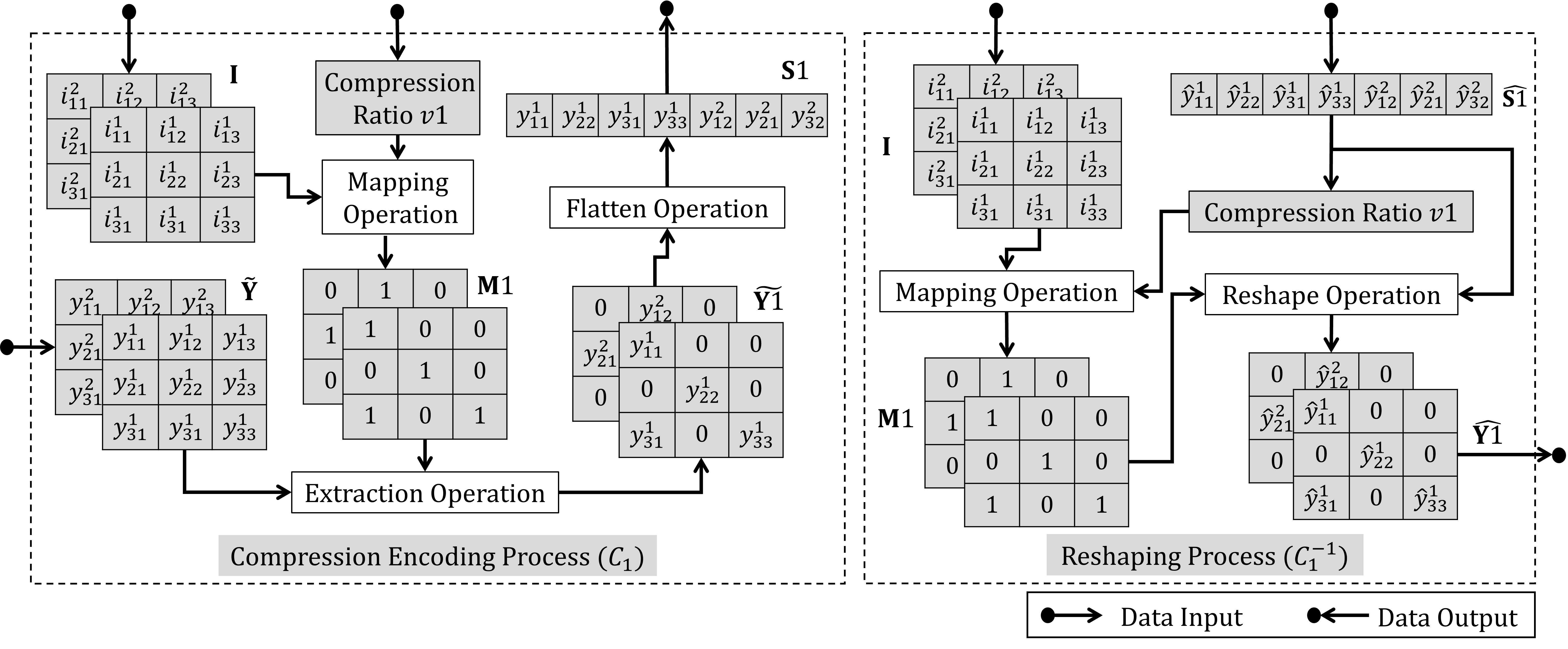}}
\caption{The compression encoding process $C_1$ based on the HEM (left) and the reshaping process $C_1^{-1}$ based on the HEM (Right). Where the compression process $C_1$ mainly selects the $(1-v1)$ proportion of features with higher importance from $\widetilde{\mathbf{Y}}$ to be transmitted over the wireless channel, and the reshaping process $C_1^{-1}$ is mainly to restore the received semantic information to their specific position in the feature matrix $\widetilde{\mathbf{Y}}$ through the importance information $\mathbf{I}$.\label{fig4}}
\end{figure*}

After obtaining the entropy model $\mathbf{P}_{\widetilde{\mathbf{Y}}|\widetilde{\boldsymbol{\sigma}}}$, the self-information $\mathbf{I}\in \mathbb{R}^{W \times H \times C2}$ of the element of feature $\mathbf{Y}$ can be obtained to measure the importance of the element of feature $\mathbf{Y}$ by the following operation:
\begin{equation}\label{eq18}
            \mathbf{I} = -\log_2 \mathbf{P}_{\widetilde{\mathbf{Y}}|\widetilde{\boldsymbol{\sigma}}}.
\end{equation}
Particularly, by employing lossless transmission indicated by the dashed arrows in Fig. \ref{fig3}, $\mathbf{I}$ is sent as the side information to the relay node $R$ and the destination node $D$, enabling them to share the entropy model with the source node $S$.

Then, according to this feature importance $\mathbf{I}$, the quantized feature $\widetilde{\mathbf{Y}}$ is effectively compressed to $\mathbf{S1}$. The specific compression process $C_1$ based on the HEM at the source node $S$ is shown in the left figure of Fig. \ref{fig4}, where the process of obtaining the mask matrix $\mathbf{M1}$ from $\mathbf{I}$ is represented as follows:
\begin{equation}\label{eq19}
            m1 = \begin{cases}
                    1, I \ge I_{S}\\
                    0, I < I_{S}.
                \end{cases}
\end{equation}

In the equation above, $m1$ and $I$ respectively represent the elements at the same position in $\mathbf{M1}$ and $ \mathbf{I}$, and $I_{S}$ is the importance threshold, which corresponds to the value of the $\left \lfloor(1-v1)L \right \rfloor$-th largest element in $ \mathbf{I}$, where $L = W \times H \times C2$ is the number of elements of $\mathbf{I}$. According to mask matrix $\mathbf{M1}$ and feature $\widetilde{\mathbf{Y}}$, sparse feature $\widetilde{\mathbf{Y1}}$ may be obtained, and then the compressed feature $\mathbf{S1}$ may also be obtained by taking out the elements in $\widetilde{\mathbf{Y1}}$ that are not zeroed. The above compression process selects the $(1-v1)$ proportion of features with higher importance from $\widetilde{\mathbf{Y}}$ to be transmitted over the wireless channel.

\subsubsection{HEC technology used at the relay node $R$}
As shown in Fig. \ref{fig3}, before the compression transformation $C_2$, it is necessary to reshape the estimation $\widehat{\mathbf{Y1}}$ of sparse semantic feature $\widetilde{\mathbf{Y1}}$ according to the importance information $\mathbf{I}$. The detailed reshaping process $C_1^{-1}$ based on the HEM is shown in the right figure of Fig. \ref{fig4}. After obtaining the compression ratio $v1$ according to the size of $\widehat{\mathbf{S1}}$, the mask matrix $\mathbf{M1}$ is obtained by Eq. (\ref{eq19}). Finally, according to $\mathbf{M1}$, the received encoding feature $\widehat{\mathbf{S1}}$ is reshaped to the sparse feature $\widehat{\mathbf{Y1}}$.

The compression transformation $C_2$ based on the HEM is to compress the reshaped sparse feature $\widehat{\mathbf{Y1}}$ according to the feature importance $\mathbf{I}$ and the compression rate $v = 1-(1-v1)(1-v2))$, where $v2 \in (0,1)$ represents the compression rate for received feature $\widehat{\mathbf{S1}}$. Then, the compressed feature $\mathbf{S2}$ will be transmitted over the $RD$ link. The detailed compression encoding process of $C_2$ is similar to the compression encoding process of $C_1$ at the source node $S$ shown in the left figure of Fig. \ref{fig4}, except that the compressed feature is $\widehat{\mathbf{Y1}}$ and the compression rate is $v$.

\subsubsection{semantic feature reshaping at the destination node $D$}
As shown in Fig. \ref{fig3}, the important information $\mathbf{I}$ is mainly used to reshape the received semantic feature $\widehat{\mathbf{S2}}$ at the destination node $D$. The detailed reshaping process $C_2^{-1}$ is similar to the reshaping process $C_1^{-1}$ at the delay node $R$ shown in the right figure of Fig. \ref{fig4}, except that the reshaping feature is $\widehat{\mathbf{S2}}$.

\subsection{Loss Function of System Model}
The optimization problem of the proposed model mainly consists of two parts: the optimization of system image reconstruction and the optimization of the HEM. Specifically, the optimization of system image reconstruction can be expressed as the MSE distortion problem of the input images $\mathbf{I}_m = \{\mathbf{I}_{m1}, \mathbf{I}_{m2}, \dots, \mathbf{I}_{mN}\}$ at the source mode $S$ and the reconstruction images $\widehat{\mathbf{I}_m} =\{\widehat{\mathbf{I}_{m1}}, \widehat{\mathbf{I}_{m2}}, \dots, \widehat{\mathbf{I}_{mN}}\}$ at the destination node $D$, which can be defined as the following E2E transmission distortion loss function:
\begin{equation}\label{eq23}
            L_{1}(\alpha_{e},\alpha_{d}) = d(\mathbf{I}_m,\widehat{\mathbf{I}_m}).
\end{equation}


The optimization problem of the HEM can be expressed as a variational autoencoder (VAE) model \cite{25}, and the goal of the inference model is to use the parametric variational density $q_{\widetilde{\mathbf{Y}},\widetilde{\mathbf{Z}}|\mathbf{S}}$ to fit the true posterior probability $p_{\widetilde{\mathbf{Y}},\widetilde{\mathbf{Z}}|\mathbf{S}}$. This goal can be optimized by minimizing the KL divergence of $p_{\widetilde{\mathbf{Y}},\widetilde{\mathbf{Z}}|\mathbf{S}}$ and $q_{\widetilde{\mathbf{Y}},\widetilde{\mathbf{Z}}|\mathbf{S}}$ over the  distribution $p_{\mathbf{S}}$ of $\mathbf{S}$ as Eq. (\ref{eq24}). The analysis of each item in the square brackets on the last line of Eq. (\ref{eq24}) is as follows.
\begin{figure*}
	\centering
    \begin{equation}\label{eq24}
        \begin{split}
             \min_{\varphi_{e}, \varphi_{h},\theta_{d}, \theta_{h}} & \mathbb{E}_{\mathbf{S} \sim p_{\mathbf{S}}}D_{KL}\Big[q_{\widetilde{\mathbf{Y}},\widetilde{\mathbf{Z}}|\mathbf{S}} \rVert p_{\widetilde{\mathbf{Y}},\widetilde{\mathbf{Z}}|\mathbf{S}} \Big]\\
             & = \min_{\varphi_{e}, \varphi_{h},\theta_{d}, \theta_{h}} \mathbb{E}_{\mathbf{S} \sim p_{\mathbf{S}}}\mathbb{E}_{\widetilde{\mathbf{Y}},\widetilde{\mathbf{Z}} \sim q_{\widetilde{\mathbf{Y}},\widetilde{\mathbf{Z}}|\mathbf{S}}} \Big[\log q_{\widetilde{\mathbf{Y}},\widetilde{\mathbf{Z}}|\mathbf{S}}(\widetilde{\mathbf{Y}},\widetilde{\mathbf{Z}}|\mathbf{S}) - \log p_{\widetilde{\mathbf{Y}},\widetilde{\mathbf{Z}}|\mathbf{S}}(\widetilde{\mathbf{Y}},\widetilde{\mathbf{Z}}|\mathbf{S}) \Big] \\
             & = \min_{\varphi_{e}, \varphi_{h},\theta_{d}, \theta_{h}} \mathbb{E}_{\mathbf{S} \sim p_{\mathbf{S}}}\mathbb{E}_{\widetilde{\mathbf{Y}},\widetilde{\mathbf{Z}} \sim q_{\widetilde{\mathbf{Y}},\widetilde{\mathbf{Z}}|\mathbf{S}}} \Big[\log q_{\widetilde{\mathbf{Y}},\widetilde{\mathbf{Z}}|\mathbf{S}}(\widetilde{\mathbf{Y}},\widetilde{\mathbf{Z}}|\mathbf{S}) - \log p_{\widetilde{\mathbf{Y}}|\widetilde{\mathbf{Z}}}(\widetilde{\mathbf{Y}}|\widetilde{\mathbf{Z}})\\
             & \quad - \log p_{\widetilde{\mathbf{Z}}}(\widetilde{\mathbf{Z}}) - \log p_{\mathbf{S}|\widetilde{\mathbf{Y}}}(\mathbf{S}|\widetilde{\mathbf{Y}}) \Big] + const,
        \end{split}
    \end{equation}
\end{figure*}

The the parametric variational density $q_{\widetilde{\mathbf{Y}},\widetilde{\mathbf{Z}}|\mathbf{S}}(\widetilde{\mathbf{Y}},\widetilde{\mathbf{Z}}|\mathbf{S})$ in the first term represents joint distribution of the hidden layer $\widetilde{\mathbf{Y}}$ and $\widetilde{\mathbf{Z}}$, which can be expressed as a joint factorized variational posterior \cite{25}:
    \begin{equation}\label{eq25}
    \begin{split}
        & q_{\widetilde{\mathbf{Y}},\widetilde{\mathbf{Z}}|\mathbf{S}}(\widetilde{\mathbf{Y}},\widetilde{\mathbf{Z}}|\mathbf{S}) \\
        & \qquad \qquad =  \prod_{i1} \mathcal{U}(\widetilde{y}_{i1}|y_{i1}-\frac{1}{2}, y_{i1}+\frac{1}{2}) \\
        & \qquad \qquad \quad \times \prod_{j1} \mathcal{U}(\widetilde{z}_{j1}|z_{j1}-\frac{1}{2}, z_{i1}+\frac{1}{2}),
    \end{split}
    \end{equation}
where $\mathcal{U}$ represents a uniform density with a width of 1, so the value of Eq. (\ref{eq25}) is 1 and the value of the first term is 0.

The second term indicates the cross-entropy of the encoding $\widetilde{\mathbf{Y}}$ and the prior (entropy model) $p_{\widetilde{\mathbf{Y}}|\widetilde{\mathbf{Z}}}(\widetilde{\mathbf{Y}}|\widetilde{\mathbf{Z}})$ can be obtained by Eq. (\ref{eq15}) and Eq. (\ref{eq16}). Furthermore, the third term indicates the cross entropy between the prior $p_{\widetilde{\mathbf{Z}}}(\widetilde{\mathbf{Z}})$ and the marginal $q_{\widetilde{\mathbf{Z}}}(\widetilde{\mathbf{Z}}) = \mathbb{E}_{\mathbf{S}\sim p_{\mathbf{S}}}\mathbb{E}_{\widetilde{\mathbf{Y}} \sim q_{\widetilde{\mathbf{Y}}|\mathbf{S}}}q_{\widetilde{\mathbf{Y}},\widetilde{\mathbf{Z}}|\mathbf{S}}(\widetilde{\mathbf{Y}},\widetilde{\mathbf{Z}}|\mathbf{S})$, and $\widetilde{\mathbf{Z}}$ can be modeled as a non-parametric fully factorized density as shown below \cite{25}:
\begin{equation}\label{eq26}
            p_{\widetilde{\mathbf{Z}}|\phi}(\widetilde{\mathbf{Z}}|\phi) = \prod_{j1} \Big(p_{z_{j1}|\phi_{j1}}(p_{z_{j1}|\phi_{j1}})*\mathcal{U}(-\frac{1}{2},\frac{1}{2}) \Big)(\widetilde{z}_{j1}),
\end{equation}
where $\phi_{j1}$ encapsulates all the parameters of $p_{z_{j1}|\phi_{j1}}$, and $*$ denotes the convolutional operation.

The fourth term represents logarithmic likelihood, which can be seen as the $\epsilon$-weighted MSE distortion term in image compression, if $p_{\mathbf{S}|\widetilde{\mathbf{Y}}}(\mathbf{S}|\widetilde{\mathbf{Y}})$ is assumed to satisfy the following distribution \cite{19,25}:
\begin{equation}\label{eq27}
            p_{\mathbf{S}|\widetilde{\mathbf{Y}}}(\mathbf{S}|\widetilde{\mathbf{Y}}) = \mathcal{N}(\mathbf{S}|\widehat{\mathbf{S}},(2\epsilon)^{-1}\mathbf{E}),
\end{equation}
where $\widehat{\mathbf{S}}$ is the output of JSCC-decoder module given in Eq. (\ref{eq9}). As shown in Fig. \ref{fig3}, $\widehat{\mathbf{Y}}$ in Eq. (\ref{eq9}) is obtained by $\widetilde{\mathbf{Y}}$ after twice compressed transmissions.

 According to the analysis of Eq. (\ref{eq24}), the loss function of the HEM shown in Fig. \ref{fig3} can be defined as follows:
\begin{equation}\label{eq28}
    \begin{split}
            L_{2} = & \mathbb{E}_{\mathbf{S} \sim p_{\mathbf{S}}}\Big[d(\mathbf{S},\widehat{\mathbf{S}}) + \\
            & \lambda \left(- \log p_{\widetilde{\mathbf{Y}}|\widetilde{\mathbf{Z}}}(\widetilde{\mathbf{Y}}|\widetilde{\mathbf{Z}})- \log p_{\widetilde{\mathbf{Z}}}(\widetilde{\mathbf{Z}})\right)\Big].
    \end{split}
\end{equation}
As can be seen from Fig. \ref{fig1}, $\mathbf{S}$ and $\widehat{\mathbf{S}}$ are the latent spatial features of system input $\mathbf{I}_m$ and reconstruction output $\widehat{\mathbf{I}_m}$, respectively. Furthermore, combining the loss function defined by Eq. (\ref{eq23}) and Eq. (\ref{eq28}), the loss function of the whole system model may be defined as Eq. (\ref{eq29}), where $\lambda$ and $\eta$ are the weight coefficients.
\begin{figure*}
	\centering
    \begin{equation}\label{eq29}
        \begin{split}
             L & = \mathbb{E}_{\mathbf{I}_{m} \sim p_{\mathbf{I}_{m}}(\mathbf{I}_{m})}\Big[\lambda \left(- \log p_{\widetilde{\mathbf{Y}}|\widetilde{\mathbf{Z}}}(\widetilde{\mathbf{Y}}|\widetilde{\mathbf{Z}})- \log p_{\widetilde{\mathbf{Z}}}(\widetilde{\mathbf{Z}}) \right) + \eta d(\mathbf{I}_m,\widehat{\mathbf{I}_m})\Big]\\
             & = \mathbb{E}_{\mathbf{I}_{m} \sim p_{\mathbf{I}_{m}}(\mathbf{I}_{m})}\Big[\lambda \Big(- \sum_{j} \log p_{\widetilde{y}_{j}|\widetilde{\sigma}_{j}}(\widetilde{y}_{j}|\widetilde{\sigma}_{j})- \sum_{j1} \log p_{\widetilde{z}_{j1}|\psi_{j1}}(\widetilde{z}_{j1}|\psi_{j1})\Big) + \eta d(\mathbf{I}_m,\widehat{\mathbf{I}_m})\Big].
        \end{split}
    \end{equation}
\end{figure*}
\subsection{Compression Parameter Optimization of System}
In order to ensure the effective operation of the system, it is assumed that the source node $S$ has the model structure of the whole system, and the source node $S$ may determine the compression ratio combination $(v1_{op}, v2_{op})$ to achieve the optimal system performance $\Phi$ according to the average fading condition of $SR$ link and $RD$ link. Then, the combination $(v1_{op}, v2_{op})$ will be sent to the compression module of the source node $S$ and the relay node $R$ as the additional information. Furthermore, the following optimization problem is formulated:
\begin{equation}\label{eq41}
        \begin{split}
              \Phi & = \max_{v1, v2\in[0,1)} \widetilde{\Phi},
        \end{split}
\end{equation}
where $\widetilde{\Phi}$ is the PSNR metric shown as follows:
\begin{equation}\label{eq46}
            PSNR(\mathbf{I}_m,\widehat{\mathbf{I}_m}) = 10\log_{10} \left(\frac{MAX_I^2}{MSE(\mathbf{I}_m,\widehat{\mathbf{I}_m})}\right),
\end{equation}
where $MAX_I$ represents the maximum possible pixel value of the input image, and for unit8 image data, $MAX_I=255$. $MSE(\mathbf{I}_m,\widehat{\mathbf{I}_m})$ represents the MSE between input image data $\mathbf{I}_m$ and reconstruction image data $\widehat{\mathbf{I}_m}$.

Since the optimization in Eq. (\ref{eq41}) depends on the MSE between input image data $\mathbf{I}_m$ and reconstruction image data $\widehat{\mathbf{I}_m}$. The optimization in Eq. (\ref{eq41}) may be regarded as a maximum match problem of compression rate combination $(v1,v2)$. Grid search algorithm is used to solve the above optimization problem, and its execution steps are as follows:
\begin{itemize}
\item Divide the search range $[0,1)$ in the direction of $v1$ and $v2$ into $K$ evenly spaced grid points, with the middle value $(v1_{k1}, v2_{k2})$ for each grid point, where $k1,k2\in\{1,2,\dots,K\}$.

\item Calculate the PSNR value obtained by the system at each grid point $(v1_{k1}, v2_{k2})$, and find out the compression ratio combination $(v1_{op}, v2_{op})$ corresponding to the optimal PSNR value.
\end{itemize}

\section{Numerical Results}
In this section, a detailed analysis of system performance will be presented in detail.
\subsection{Experiments Setup}
\subsubsection{System and Model Parameters Setup}
In all simulations, the parameter settings of the system and the model training parameters are shown in Table \ref{tab:TABLE II}. The detailed network structure parameters of the system are shown in Table \ref{tab:TABLE III}. In the first column, below each model module name are the input and output sizes of the modules. The parameters in brackets in the layer structure column represents parameters $( in-channels, out-channels, kernel-size, stride, padding, output-padding )$. In the third column are the activation function corresponding to the network layer in the second column.
\begin{table}[]
\centering
\caption{System simulation and model training parameters}
\label{tab:TABLE II}
\renewcommand{\arraystretch}{1.3}
\begin{tabular}{cc}
\toprule [1pt]
System Simulation Parameters                    & Value         \\
\midrule
$SR$ link noise power $N_R$   & -80 dBm   \\
$RD$ link noise power $N_D$   & -80 dBm   \\
The distance $d_{SD}$  between $S$ and $D$      & 100 m      \\
Path-loss parameter $a$                        & 3          \\
Training rounds $Ep$                                           & 20            \\
Number of input iamge $N$                       & 2            \\
shared information extraction rate $\gamma_{p}$ & 0.5            \\
Optimizer                                       & Adam          \\
Learning rate $\xi$                             & 0.0001        \\
Loss function factor  $\lambda$                 & 8192          \\
Loss function factor $\eta$                     & $\frac{1}{3\times512\times1024}$  \\
\bottomrule [1pt]
\end{tabular}
\end{table}

\begin{table}[]
\centering
\caption{System network structure parameters}
\label{tab:TABLE III}
\renewcommand{\arraystretch}{1.3}
\begin{tabular}{|c|c|c|}
\hline
Module & Layer Structure & Activation \\ \hline
\multirow{4}{*}{\begin{tabular}[c]{@{}c@{}}Latent Transform $LT_e$\\ 3x512x1024-64x64x128\end{tabular}} & Conv2d (3,64,3,1,1)  & GDN \\ \cline{2-3}
 & Conv2d (64,128,3,2,1)  & GDN       \\ \cline{2-3}
 & Conv2d (128,256,3,2,1)  & GDN        \\ \cline{2-3}
 & Conv2d (256,64,3,2,1)  & None        \\ \hline
\multirow{2}{*}{\begin{tabular}[c]{@{}c@{}}JSCC Encoder $A_e$\\ 96x64x128-96x64x128\end{tabular}} & Conv2d (96,48,3,1,1)  & GDN        \\ \cline{2-3}
                                                                                    & Conv2d (48,96,3,1,1)  & None       \\ \hline
\multirow{3}{*}{\begin{tabular}[c]{@{}c@{}}nonlinear transformation $h_a$\\ 96x64x128-32x16x32\end{tabular}}  & Conv2d (96,32,3,1,1)  & Relu       \\ \cline{2-3}
                                                                                    & Conv2d (32,32,5,2,2)  & Relu       \\ \cline{2-3}
                                                                                    & Conv2d (32,32,5,2,2)  & None       \\ \hline
\multirow{3}{*}{\begin{tabular}[c]{@{}c@{}}nonlinear transformation $h_s$\\ 32x16x32-96x64x128\end{tabular}} & ConvT (32,32,5,2,2,1) & Relu \\ \cline{2-3}
                                                                                    & ConvT (32,32,5,2,2,1) & Relu       \\ \cline{2-3}
                                                                                    & ConvT (32,96,3,1,1,0)   & Relu       \\ \hline
\multirow{2}{*}{\begin{tabular}[c]{@{}c@{}}JSCC Decoder $A_d$\\ 96x64x128-96x64x128\end{tabular}} & Conv2d (96,48,3,1,1)  & GDN        \\ \cline{2-3}
                                                                                    & Conv2d (48,96,3,1,1)  & None       \\ \hline
\multirow{4}{*}{\begin{tabular}[c]{@{}c@{}}Latent Inversion $LT_d$\\ 64x64x128-3x512x1024\end{tabular}} & ConvT (64,256,3,2,1,1) & GDN \\ \cline{2-3}
 & ConvT (256,128,3,2,1,1) & GDN        \\ \cline{2-3}
 & ConvT (128,64,3,2,1,1) & GDN        \\ \cline{2-3}
 & Conv2d (64,3,3,1,1)  & Tanh       \\ \hline
\end{tabular}
\end{table}

\subsubsection{Model training Details}
Specially, the model was trained and tested using the Cityscapes data set with image size of $3\times2048\times1024$. Prior to being fed into the model, the images in the data set were down-sampled to $3\times512\times1024$ and normalized to the interval of $[0,1]$. The training process of the system model is shown in Alg. \ref{Alg.1}, and the training of the whole model is completed based on dual RTX A6000 GPU.

\begin{algorithm}
    \caption{Training the System Model}
    \label{Alg.1}
    \begin{algorithmic}[1]
        \Require Training data $\mathbf{I}_{m}$, the loss function factors $\lambda$ and $\eta$, learning rate $\xi$, training rounds $Ep$, path-loss factor $a$, shared information extraction rate $\gamma_{p}$, compression ratio $v1=0$ and $v2=0$, noise power $N_R = N_D = -66$ dBm and total transmitted power $P = 0$ dBm, distance $d_{SR} = d_{RD} = 1 m$.
            \State Randomly initialize $k = 1$ and model parameters $W^{(0)} = \{\alpha_{e}^{(0)}, \varphi_{e}^{(0)}, \varphi_{h}^{(0)},\theta_{d}^{(0)}, \theta_{h}^{(0)}, \alpha_{d}^{(0)}\}$.
            \While{$k \leq Ep$}
                \State Input data $\mathbf{I}_{m}$ downsampling,
                \State Calculate loss function  $L\left(W^{(k-1)}, \lambda, \eta \right)$ according to Eq. (\ref{eq29}),
                \State Calculate the gradients $\nabla_{W^{(k-1)}}L\left(W^{(k-1)}, \lambda, \eta \right) $,
                \State Update $W^{(k)} \leftarrow W^{(k-1)}- \xi\nabla_{W^{(k-1)}}L\left(W^{(k-1)}, \lambda, \eta \right)$
                \State $k=k+1$,
                \If {$k > Ep $}
                \State break,
                \EndIf
            \EndWhile
            \Ensure Trained model parameters \\ $W^{(Ep)} = \{\alpha_{e}^{(Ep)}, \varphi_{e}^{(Ep)}, \varphi_{h}^{(Ep)},\theta_{d}^{(Ep)}, \theta_{h}^{(Ep)}, \alpha_{d}^{(Ep)}\}$.
    \end{algorithmic}
\end{algorithm}

\subsubsection{Evaluation Metrics and Comparison Schemes}
The PSNR metric and MS-SSIM metric for image transmission were used to verify the image semantic transmission performance of the proposed system. Specially, the formula for calculating PSNR is as shown in Eq. (\ref{eq46}), and the formula for calculating MS-SSIM \cite{26} is as follows:
\begin{equation}\label{eq47}
\begin{split}
            & MS-SSIM(\mathbf{I}_m,\widehat{\mathbf{I}_m}) = \\
            & \quad [l_M(\mathbf{I}_m,\widehat{\mathbf{I}_m})]^{\alpha_{M}}\prod_{j=1}^M [c_j(\mathbf{I}_m,\widehat{\mathbf{I}_m})]^{\beta_{j}}[s_j(\mathbf{I}_m,\widehat{\mathbf{I}_m})]^{\gamma_{j}}.
\end{split}
\end{equation}
where $l(\mathbf{I}_m,\widehat{\mathbf{I}_m})$, $c(\mathbf{I}_m,\widehat{\mathbf{I}_m})$ and $s(\mathbf{I}_m,\widehat{\mathbf{I}_m})$ denote luminance, contrast and structure comparison measures,  respectively. Exponents $\alpha_{M}$, $\beta_{j}$ and $\gamma_{j}$ are used to adjust the relative importance of different components.

Three different transmission schemes are used as comparison schemes to verify the performance of the proposed transmission scheme. The first is the scheme that replaces the shared features extraction technology in the proposed system with the one proposed in \cite{18}. The second is the scheme that uses only HEC technology in the proposed system. Additionally, the third is the LSCI scheme proposed in \cite{6}. In order to simplify the representation, the above three comparison schemes and the scheme proposed in this paper are represented as ED-HEM, HEM, LSCI and PC-HEM, respectively.

\subsection{Result Analysis}
\begin{figure*}[htbp]
	\centering
	\begin{subfigure}{0.44\linewidth}
		\centering
		\includegraphics[width=1.1\linewidth]{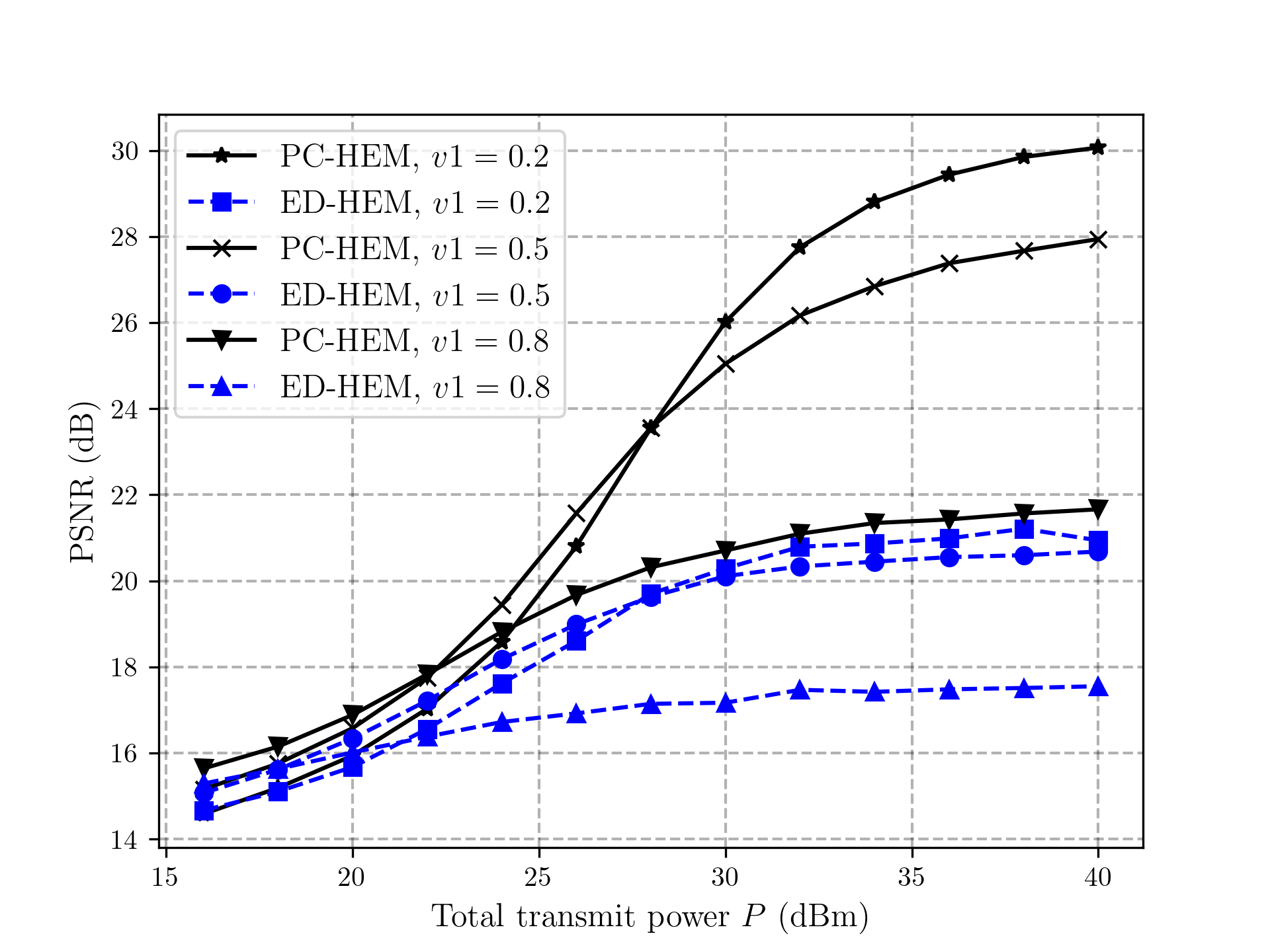}
		\caption{PSNR}
		\label{(a)}
	\end{subfigure}
    \centering
	\begin{subfigure}{0.44\linewidth}
		\centering
		\includegraphics[width=1.1\linewidth]{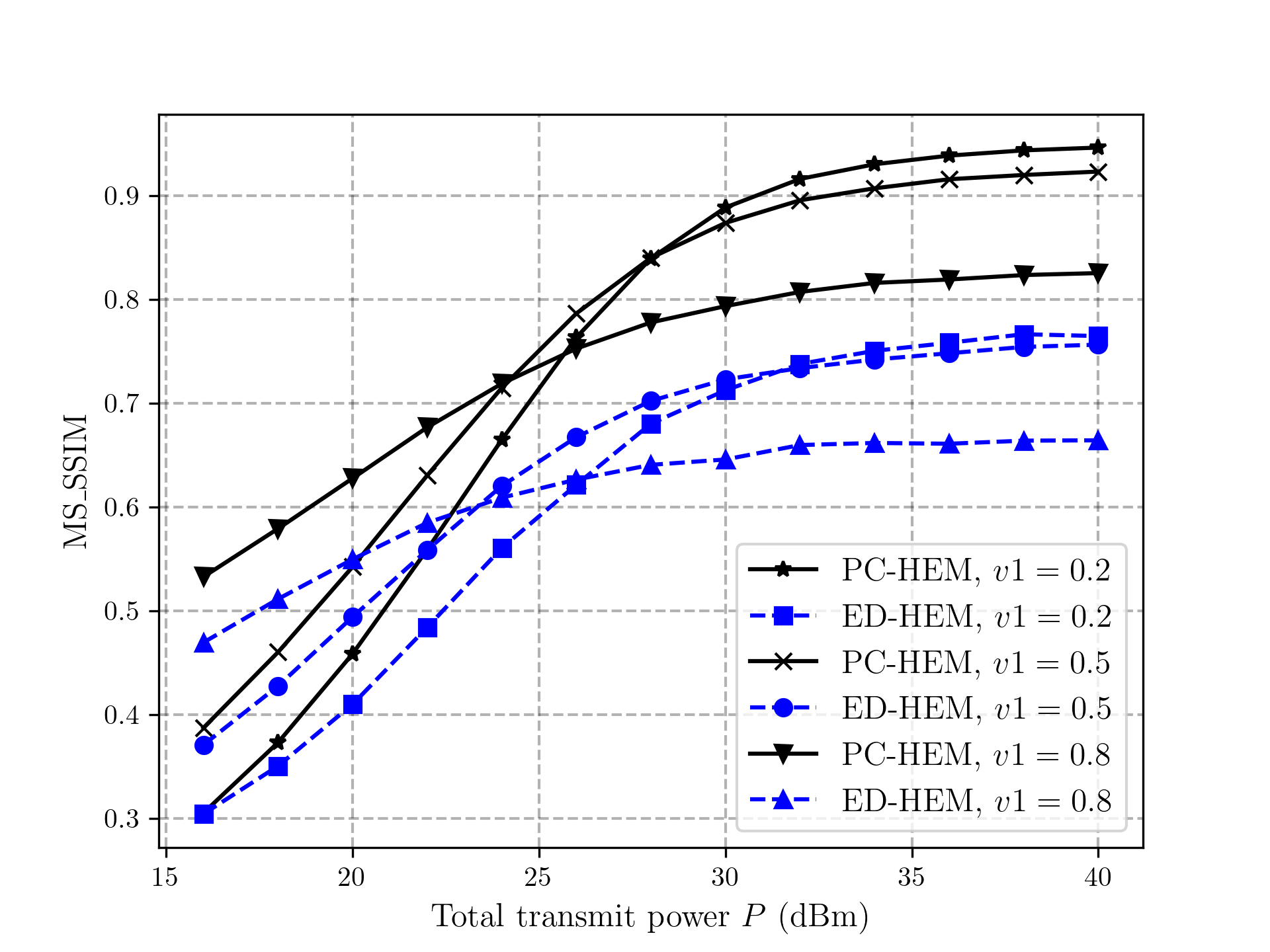}
		\caption{MS-SSIM}
		\label{(b)}
	\end{subfigure}
	\caption{PSNR and MS-SSIM of system against the transmit power $P$ for PC-HEM and ED-HEM schemes, with parameters $v2=0.2$, $d_{SR} = d_{RD} = 50$ m, and three different compression ratio ($v1$ = 0.2, 0.5 and 0.8).}
	\label{fig5}
\end{figure*}

\begin{figure*}[htbp]
	\centering
	\begin{subfigure}{0.44\linewidth}
		\centering
		\includegraphics[width=1.1\linewidth]{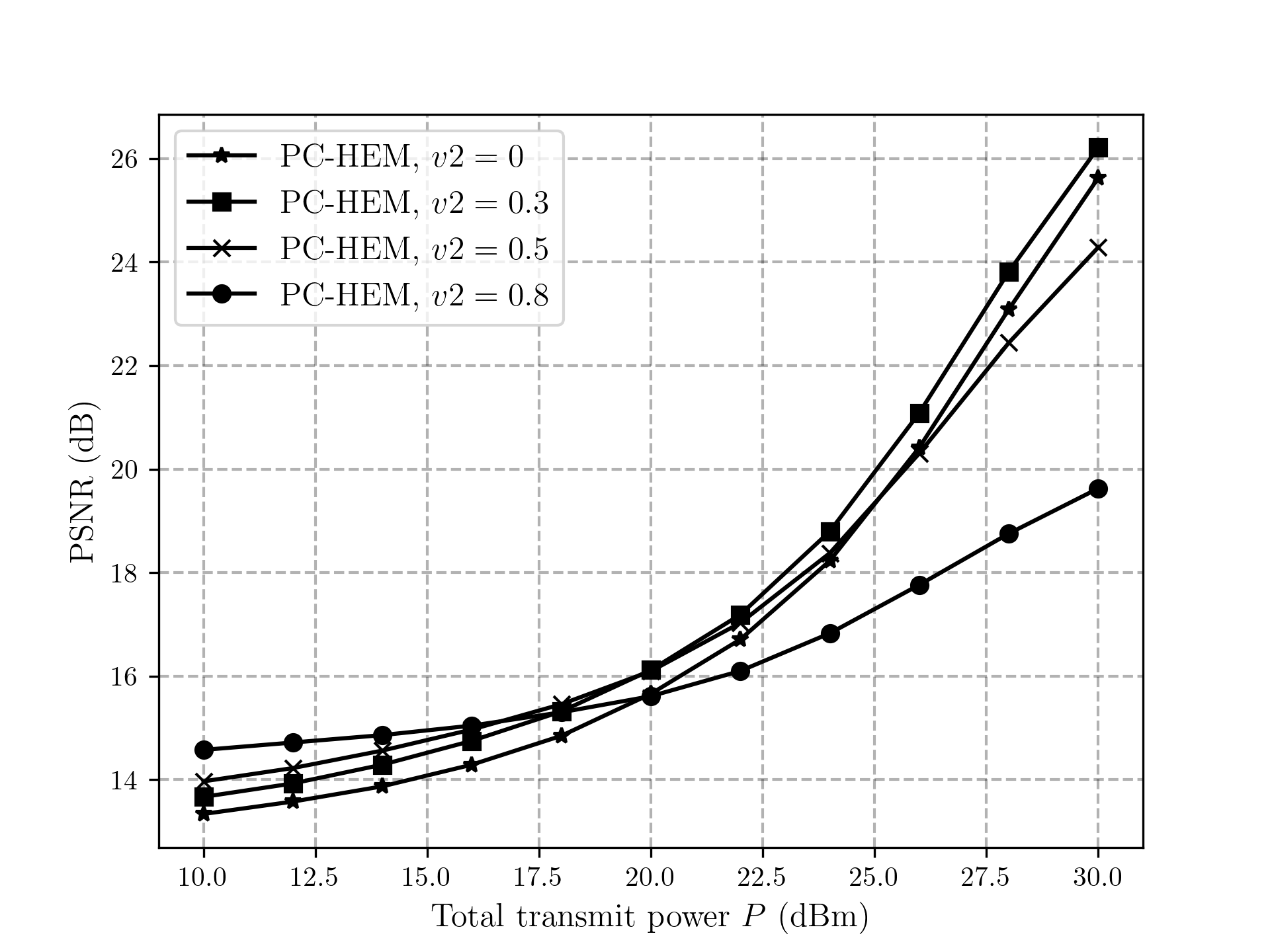}
		\caption{PSNR}
		\label{a}
	\end{subfigure}
    \centering
	\begin{subfigure}{0.44\linewidth}
		\centering
		\includegraphics[width=1.1\linewidth]{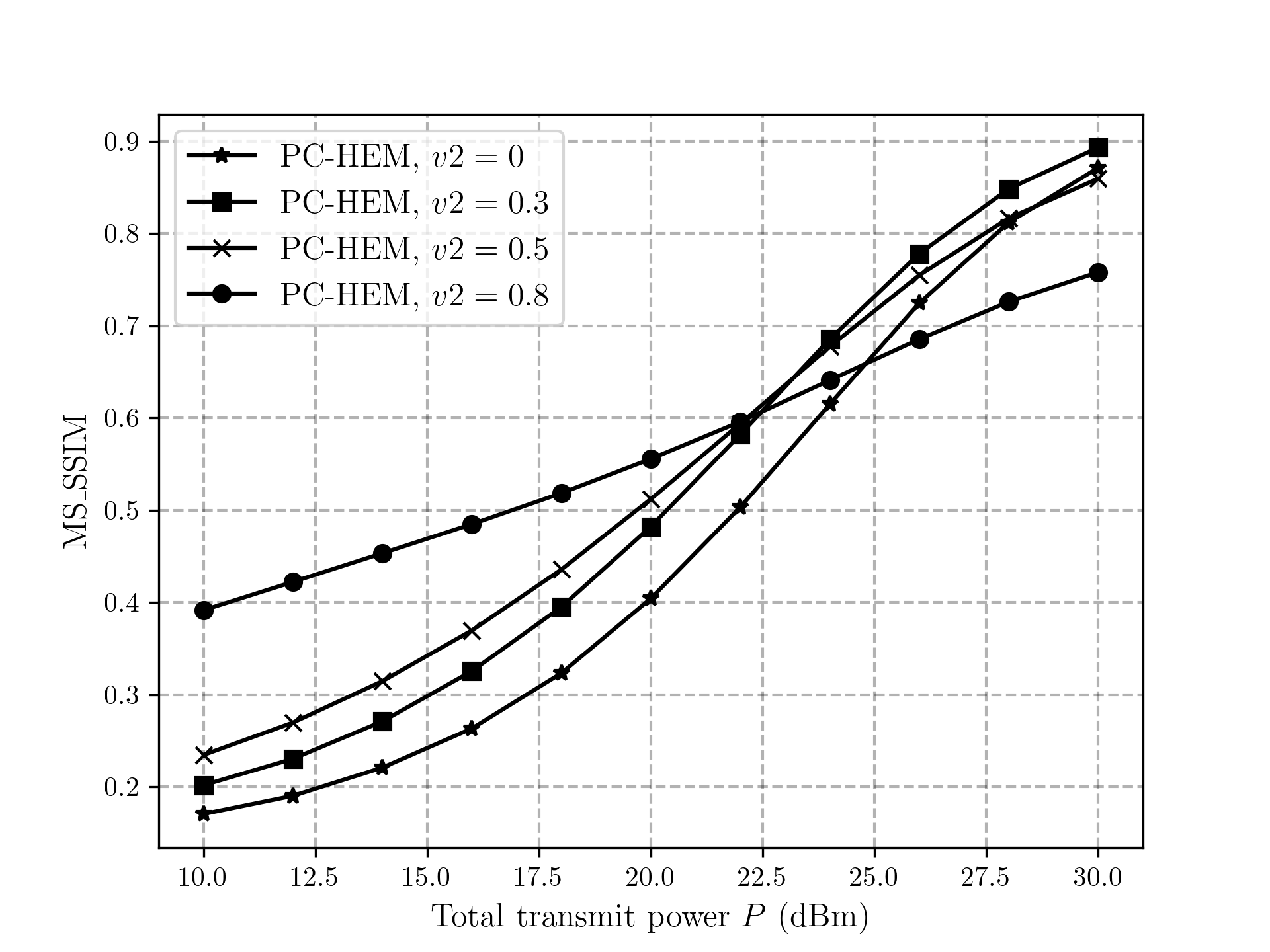}
		\caption{MS-SSIM}
		\label{b}
	\end{subfigure}
	\caption{PSNR and MS-SSIM of system against the transmit power $P$ for PC-HEM scheme, with parameters $v1=0.2$, $d_{SR} = d_{RD} = 50$ m, and four different compression ratio ($v2$ = 0, 0.2, 0.5 and 0.8).}
	\label{fig6}
\end{figure*}

Fig. \ref{fig5} illustrates the variation of PSNR and MS-SSIM with respect to the transmit power $P$ for the PC-HEM and ED-HEM schemes, considering three different compression ratios ($v1$ = 0.2, 0.5, and 0.8). From Fig. \ref{fig5}, it is easily observed that the PC-HEM scheme outperforms the ED-HEM scheme in terms of both PSNR and MS-SSIM. Specifically, at $P=40$ dBm and $v1=0.2$, the PC-HEM scheme exhibits an approximate 9 dB advantage in PSNR and a approximate 0.2 advantage in MS-SSIM compared to the ED-HEM scheme. Additionally, as $P$ decreases, larger values of $v1$ result in better PSNR and MS-SSIM performance, while for larger $P$ values, larger $v1$ values lead to poorer PSNR and MS-SSIM performance. This is because at lower $P$ values, the system performance is heavily influenced by the average SNR of the transmitted semantic data, and larger $v1$ values result in a higher average SNR. Conversely, at higher P values, the system performance is primarily affected by the amount of transmitted semantic data, and larger $v1$ values lead to a smaller amount of transmitted semantic data.

Fig. \ref{fig6} presents the variation of PSNR and MS-SSIM with respect to the transmit power $P$ for the PC-HEM scheme, considering four different compression ratio combinations ($v2$ = 0, 0.2, 0.5 and 0.8). As can be seen from Fig. \ref{fig6}, when the transmit power $P$ is low, the larger the compression ratio $v2$ of the received semantic data at the relay node $R$, the better the PSNR and MS-SSIM performance of the system. Specifically, when $P$ = 10 dBm, compared to $v2$ = 0, $v2$ = 0.8 results in an increase of approximately 1.3 dB and 0.22 in PSNR and MS-SSIM, respectively. However, as $P$ increases, the larger the value of $v2$, the slower the increase in PSNR and MS-SSIM. In particular, when $P$ = 30 dBm, compared to $v2$ = 0.8, $v2$ = 0 leads to an increase of approximately 6 dB in PSNR and an increase of approximately 0.11 in MS-SSIM. This phenomenon may be attributed to the fact that at lower $P$ values, a higher compression ratio $v2$ ensures a higher average transmit power $\bar{P}$ for semantic feature data over the $RD$ link, thereby guaranteeing the effective transmission of the more important semantic features. However, when $P$ is larger, the smaller the compression ratio $v2$ is, the more semantic information will be effectively transmitted by $RD$ link.

\begin{figure*}[htbp]
	\centering
	\begin{subfigure}{0.44\linewidth}
		\centering
		\includegraphics[width=1.1\linewidth]{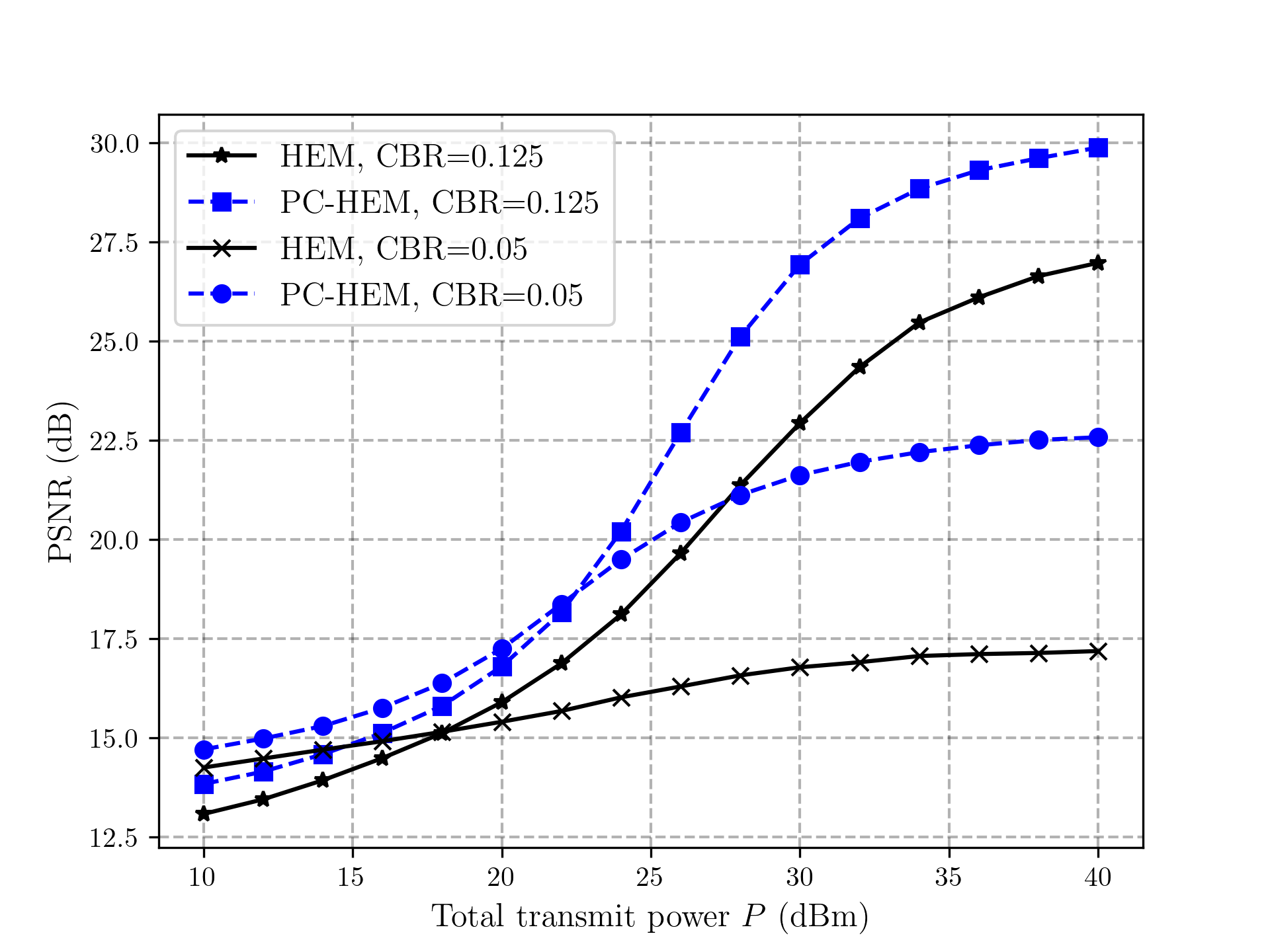}
		\caption{PSNR}
		\label{a}
	\end{subfigure}
    \centering
	\begin{subfigure}{0.44\linewidth}
		\centering
		\includegraphics[width=1.1\linewidth]{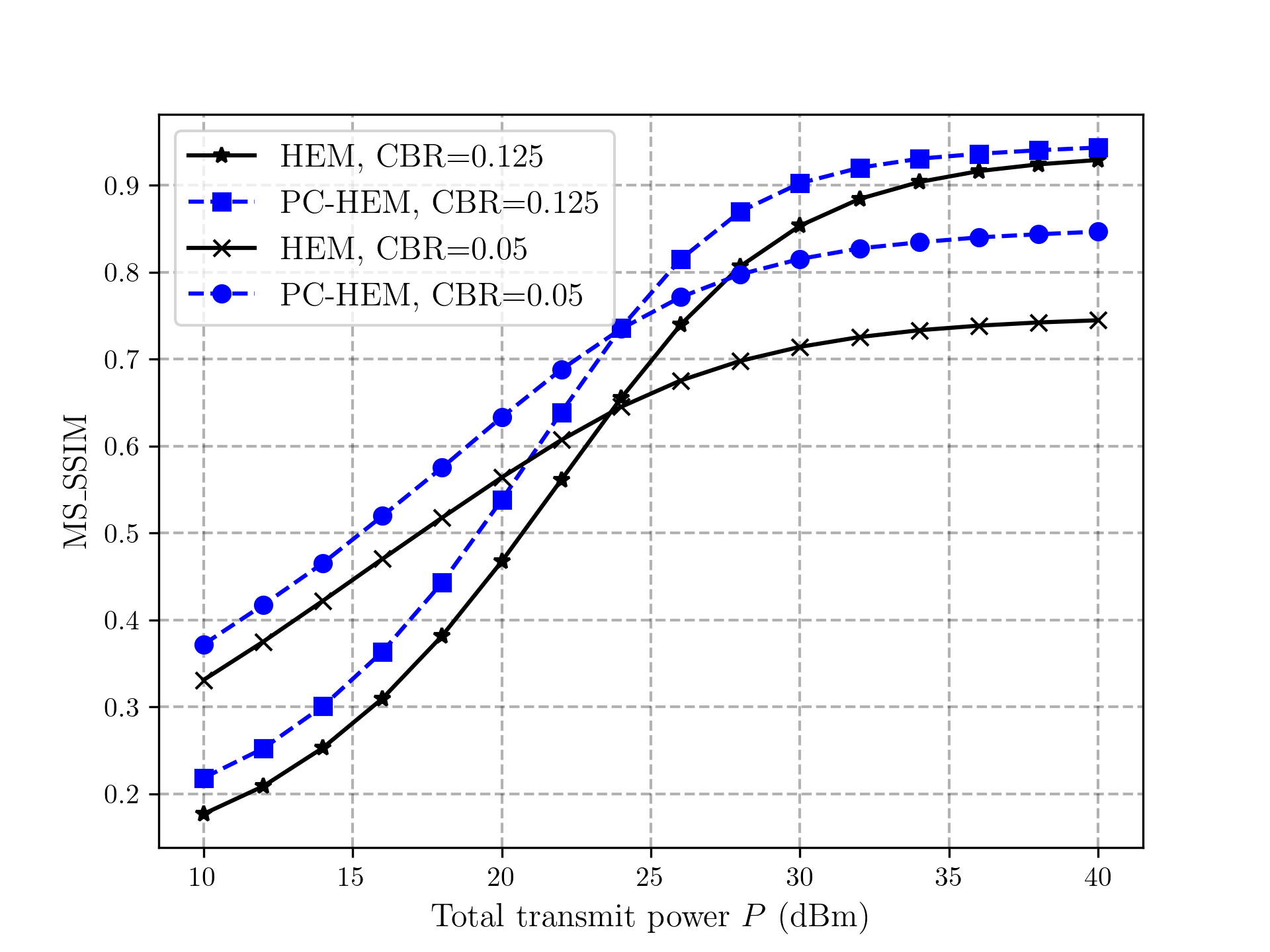}
		\caption{MS-SSIM}
		\label{b}
	\end{subfigure}
	\caption{PSNR and MS-SSIM of system against the transmit power $P$ for PC-HEM and HEM schemes, with parameters $v2=0$,  $d_{SR} = d_{RD} = 50$ m, and two different channel bandwidth ratio CBR = 0.125 and 0.05 at the source node $S$.}
	\label{fig7}
\end{figure*}

Fig. \ref{fig7} demonstrates the variation of PSNR and MS-SSIM with respect to the transmit power $P$ for the PC-HEM and HEM schemes, considering two different CBR = 0.125 and 0.05 of the source node $S$. From Fig. \ref{fig7}, it is evident that the PC-HEM scheme achieves better PSNR and MS-SSIM performance compared to the HEM scheme. Specifically, at $P = 40$ dBm and CBR = 0.05, the PC-HEM scheme exhibits an approximate 5 dB advantage in PSNR and an approximate 0.1 advantage in MS-SSIM over the HEM scheme. This is because the adopted shared features extraction technology based on Pearson correlation effectively reduces the dimension of the semantic latent feature space, thereby improving the efficiency of semantic feature encoding and transmission. Furthermore, at lower values of $P$, a smaller CBR corresponds to better PSNR and MS-SSIM performance, while at higher values of $P$, a larger CBR leads to better PSNR and MS-SSIM performance. The underlying reasons are consistent with the analysis of similar phenomena discussed in Fig. \ref{fig5}.

\begin{figure}
\centerline{\includegraphics[width=3.5in]{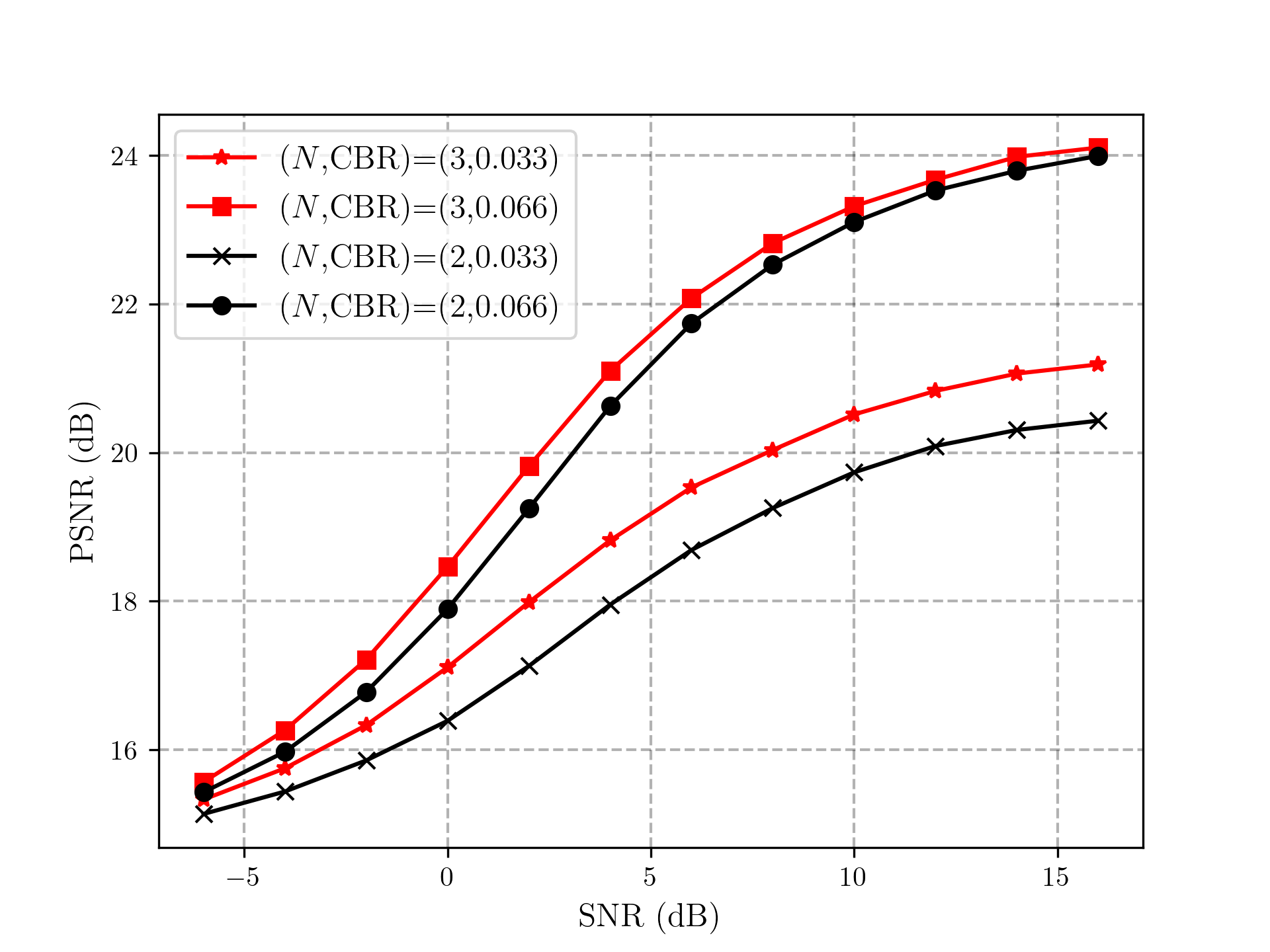}}
\caption{PSNR of system against the SNR for PC-HEM scheme, with parameters $v2=0$, $d_{SR} = d_{RD} = 50$ m, $N \in \{2, 3\}$ and two different channel bandwidth ratio CBR = 0.033 and 0.066 at the source node $S$. \label{fig8}}
\end{figure}

Fig. \ref{fig8} depicts the variation of PSNR with respect to SNR for the PC-HEM scheme, considering two different values for $N$ ($N$ = 2 and 3) and two different CBR values (CBR = 0.033 and 0.066) at the source node $S$. The results from Fig. \ref{fig8} clearly indicate that the PC-HEM scheme with $N = 3$ achieves superior PSNR performance compared to the PC-HEM scheme with $N = 2$. Specifically, at SNR = $2$ dB and CBR = 0.033, the PC-HEM scheme with $N = 3$ exhibits an approximate 0.9 dB advantage in PSNR over the PC-HEM scheme with $N = 2$. This is because in the case of the same shared feature extraction rate $\gamma_{p}$ and the CBR, the $N=3$ system has a larger latent feature space compression rate and a smaller hyperprior entropy compression rate $v1$ than the $N=2$ system, thereby transmitting more important semantic features.

\begin{figure*}[htbp]
	\centering
	\begin{subfigure}{0.44\linewidth}
		\centering
		\includegraphics[width=1.1\linewidth]{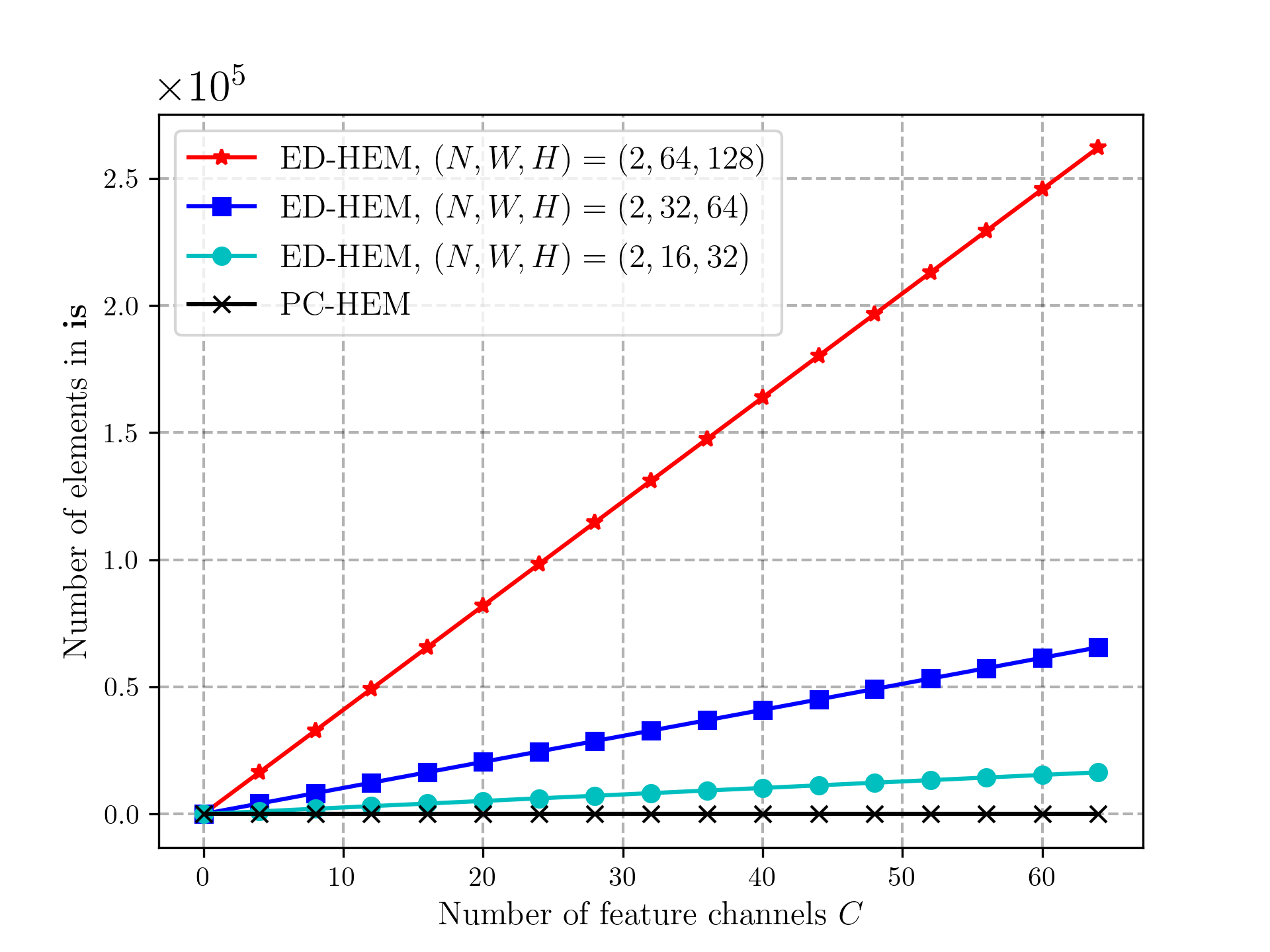}
		\caption{Number of elements in $\mathbf{is}$}
		\label{a}
	\end{subfigure}
    \centering
	\begin{subfigure}{0.44\linewidth}
		\centering
		\includegraphics[width=1.1\linewidth]{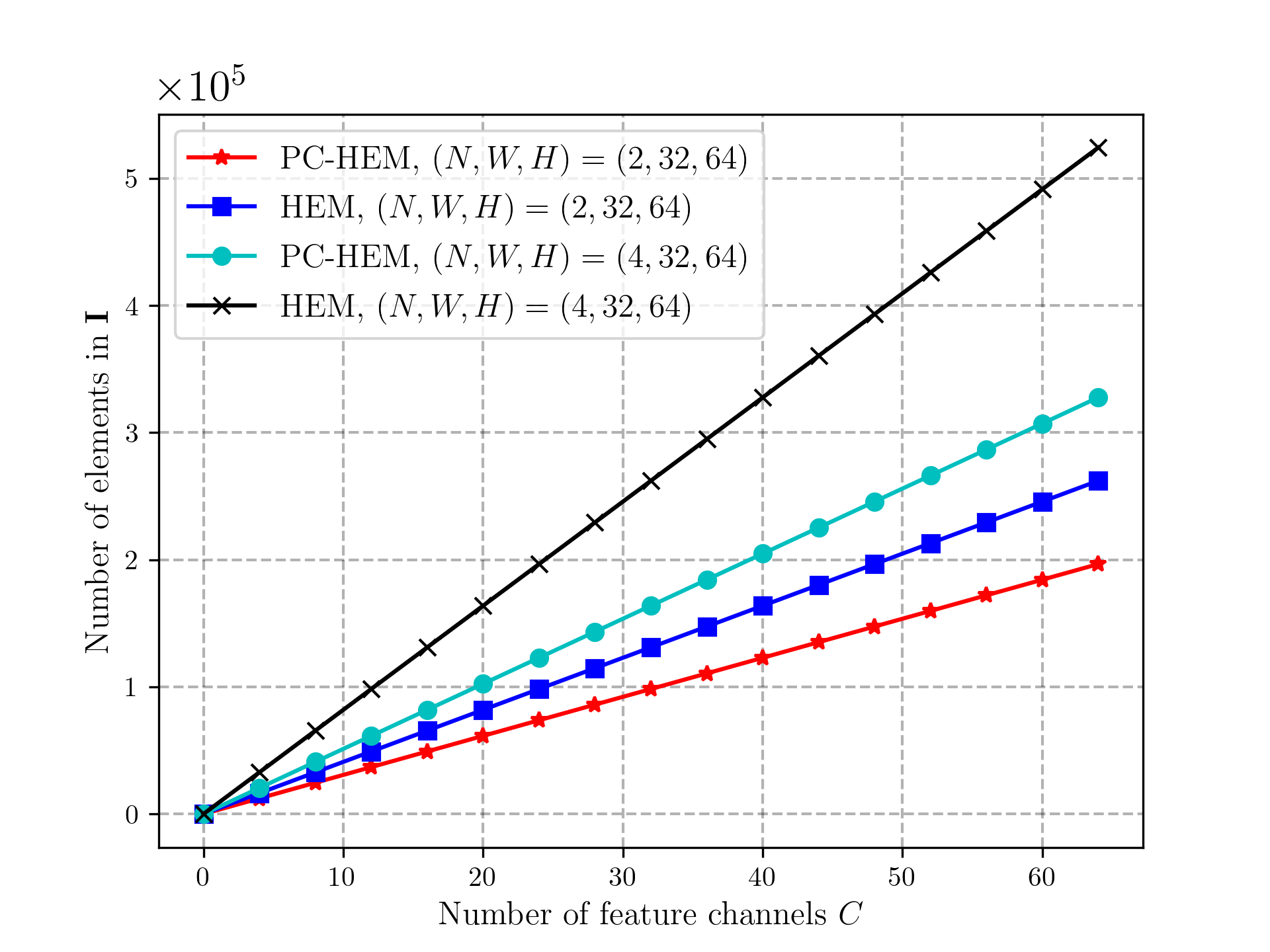}
		\caption{Number of elements in $\mathbf{I}$}
		\label{b}
	\end{subfigure}
	\caption{Additional transmission overhead required by the PC-HEM scheme. (a) compares the number of elements in the shared channel index vector $\mathbf{is}$ that need to be transmitted for PC-HEM and ED-HEM schemes with $N=2$ input images. (b) compares the number of elements in the important information matrix $\mathbf{I}$ that need to be transmitted for PC-HEM and HEM schemes with $N \in \{2, 4\}$ input images.}
	\label{fig9}
\end{figure*}

Fig. \ref{fig9} illustrates the additional transmission overhead required by the PC-HEM scheme. Fig. \ref{fig9}\subref{a} shows the relationship between the number of elements in shared channel index vector $\mathbf{is}$ that need to be transmitted and the number $C$ of latent transform output channels. The comparison is made between the PC-HEM scheme and ED-HEM scheme with $N=2$ input images, considering three different channel feature sizes $(W,H)$ = (64,128),(32,64) and (16,32). From Fig. \ref{fig9}\subref{a}, it can be observed that the number of elements in shared channel index vector $\mathbf{is}$ is zero for the PC-HEM scheme. This is because the PC-HEM scheme does not require the transmission of shared channel indexes. In contrast, the ED-HEM scheme requires an increasing number of elements in the shared channel index vector $\mathbf{is}$ as $C$ and $(W,H)$ increase. Specifically, when $C=60$ and $(W,H)$ = (64,128), the required number of elements for the ED-HEM scheme is approximately $2.5\times10^5$.

Furthermore, Fig. \ref{fig9}\subref{b} shows the variation of the number of elements in the importance matrix $\mathbf{I}$ against the number $C$ of latent transform output channels for the PC-HEM and HEM schemes, considering $(W,H)$ = (32,64) and $N \in \{2, 4\}$. It can be observed from Fig. \ref{fig9}\subref{b} that the PC-HEM scheme requires fewer elements in the importance matrix $\mathbf{I}$ compared to the ED-HEM scheme. Specifically, when $C = 60$ and $N = 4$, the number of elements in the importance matrix $\mathbf{I}$ for the HEM scheme is $5\times10^5$, while for the PC-HEM scheme, it is $3\times10^5$. According to the analysis of Fig. \ref{fig9}, it is concluded that the proposed PC-HEM scheme requires lower additional information transmission overhead compared to the ED-HEM and HEM schemes.

\begin{figure}
\centerline{\includegraphics[width=3.5in]{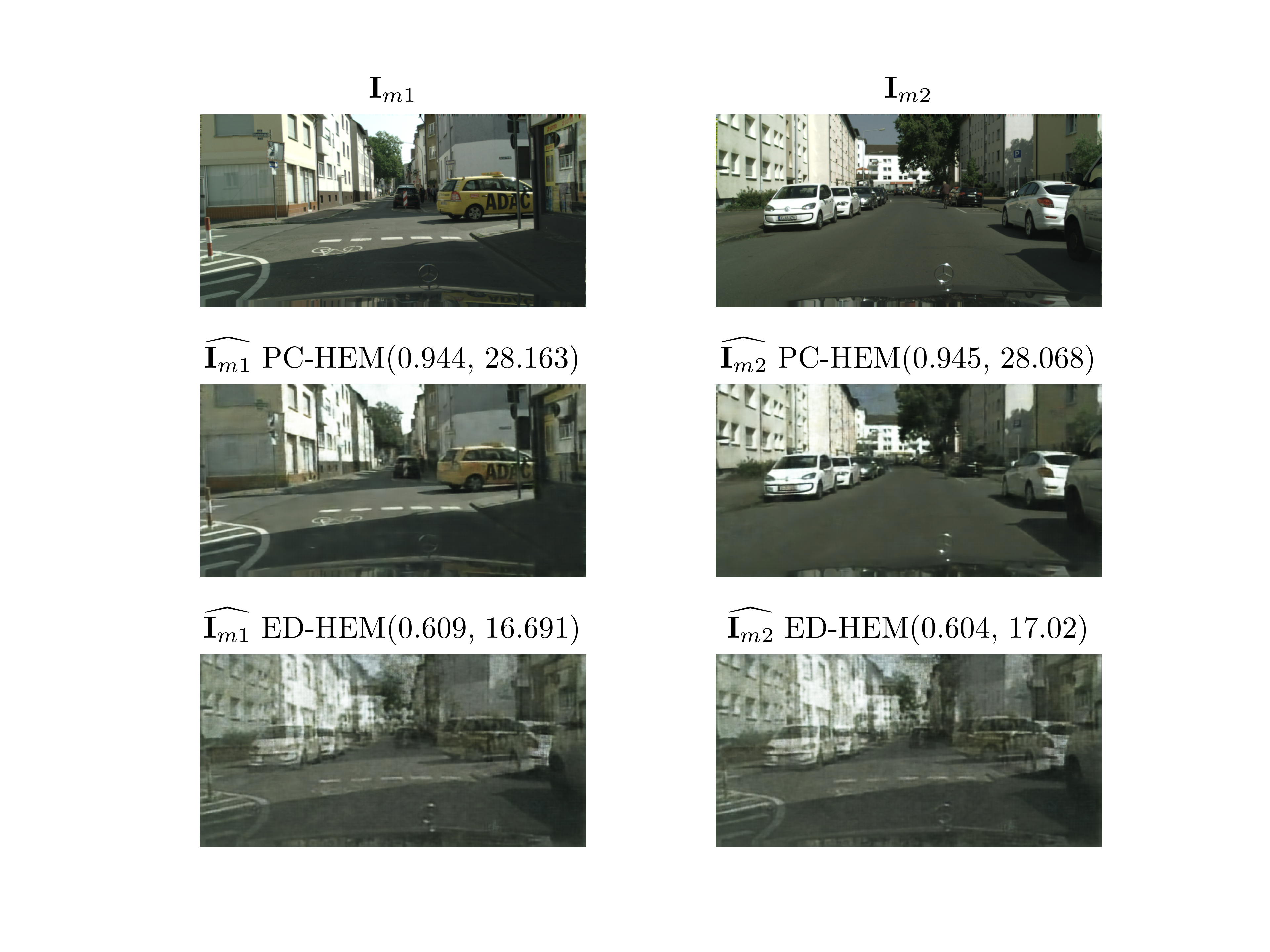}}
\caption{Comparison of image recovery effect between PC-HEM and ED-HEM schemes, with parameters $v1=0.4$, $v2=0.1$, $P = 35$ dBm, $d_{SR}=d_{RD} = 50$ m. where, the numbers in parentheses indicate PSNR and MS-SSIM, respectively. \label{fig10}}
\end{figure}

Fig. \ref{fig10} illustrates the comparison of image recovery effect between PC-HEM and ED-HEM schemes. Where, the top, middle and bottom are the two input images $\mathbf{I}_{m1}$ and $\mathbf{I}_{m2}$ at the source node $S$, the recovered images $\widehat{\mathbf{I}_{m1}}$ and $\widehat{\mathbf{I}_{m2}}$ at the destination node $D$ of PC-HEM and ED-HEM schemes, respectively. In addition, the data in parentheses represents the MS-SSIM and PSNR performance of the system, respectively. It can be clearly seen from the restored images and performance data in Fig. \ref{fig10} that the image recovery effect of PC-HEM scheme is  better than that of ED-HEM scheme. In particular, the MS-SSIM performance of the PC-HEM scheme is approximate 0.34 better than that of the ED-HEM scheme, and the PSNR performance of the PC-HEM scheme is about 11 dB better than that of the ED-HEM scheme.

\begin{figure}
\centerline{\includegraphics[width=3.5in]{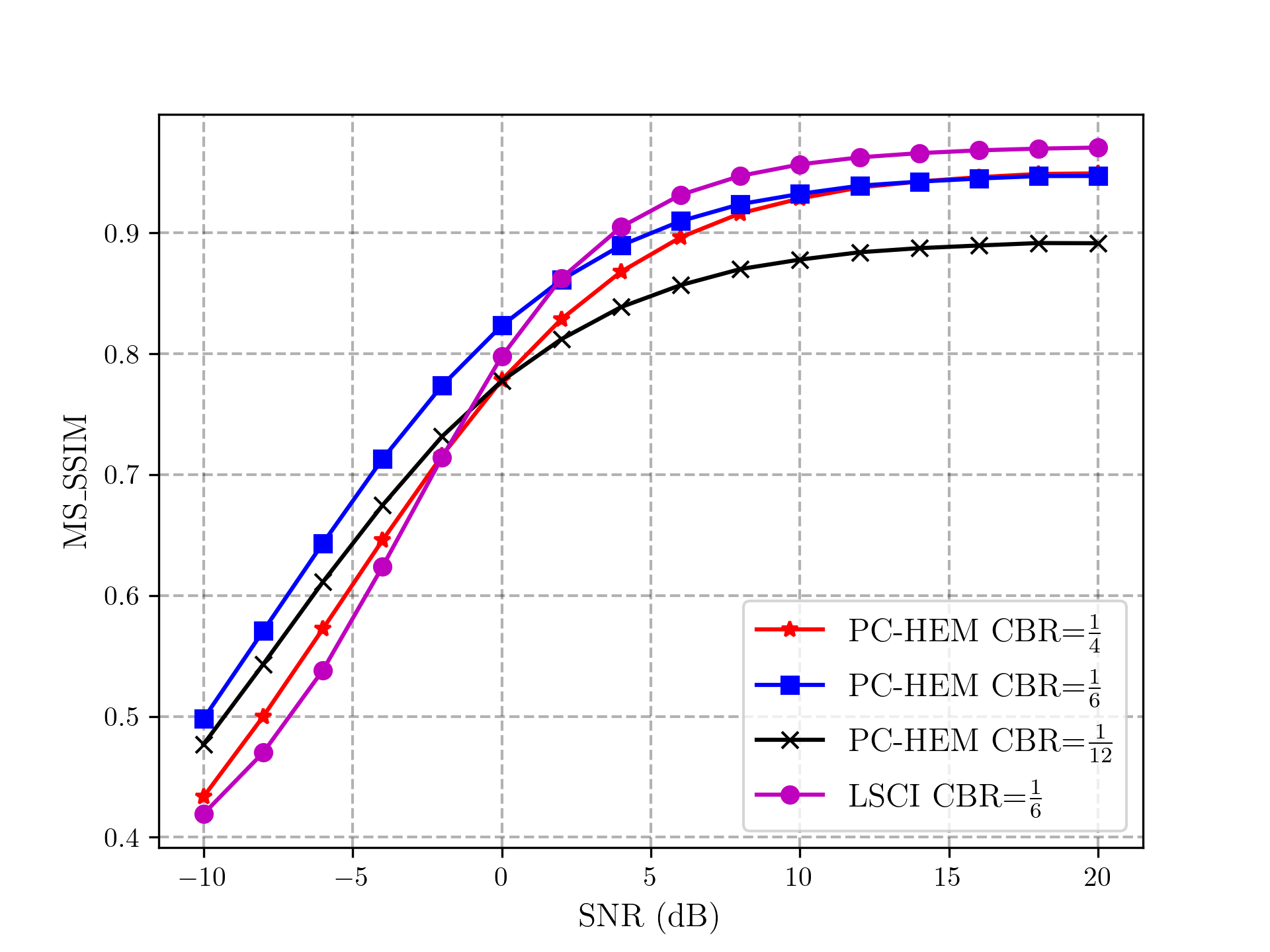}}
\caption{MS-SSIM of the E2E communication system between the source node $S$ and the destination node $D$ against the SNR for PC-HEM and LSCI schemes, with parameters  $d_{SD} = 100$ m and three different channel bandwidth ratio CBR = $\frac{1}{4}$, $\frac{1}{6}$ and $\frac{1}{12}$ at the source node $S$. \label{fig11}}
\end{figure}

Fig. \ref{fig11} depicts the variation of MS-SSIM of the E2E communication system between the source node $S$ and the destination node $D$ respect to SNR for the PC-HEM and LSCI schemes, considering three different CBR values (CBR = $\frac{1}{4}$, $\frac{1}{6}$ and $\frac{1}{12}$) at the source node $S$. From Fig. \ref{fig11}, it is evident that in the case of lower SNR, the PC-HEM scheme achieves better MS-SSIM performance compared to the LSCI scheme. Specifically, at SNR = $-5$ dB and CBR = $\frac{1}{6}$, the PC-HEM scheme exhibits an approximate 0.1 advantage in MS-SSIM over the LSCI scheme. This is because the shared features extraction technology based on Pearson correlation and the HEC technology make the PC-HEM scheme more effective than the LSCI scheme in extracting important features of semantic information.

\section{Conclusion}
This paper proposes a semantic image transmission relay communication network based on shared feature extraction and hyperprior entropy compression. Specifically, shared feature extraction technology based on Pearson correlation is used to reduce redundancy among the semantic latent features of input images. Moreover, a hyperprior entropy compression technology is used to efficiently compress transmission data, according to the conditions of channel noise and link fading. The experiment results show that compared to recent research methods, the proposed system exhibits lower additional transmission overhead and achieves higher PSNR and MS-SSIM performance for semantic image transmission. Under identical conditions, the system exhibits an approximately 0.2 higher MS-SSIM compared to the comparative method. Building upon the research on fixed-ratio shared feature extraction presented in this paper, an adaptive shared feature extraction scheme emerges as a promising direction for further exploration.

\end{document}